%% file: acl_latex.tex
\documentclass[11pt]{article}

\usepackage[preprint]{acl}

\usepackage{times}
\usepackage{latexsym}

\usepackage[T1]{fontenc}

\usepackage[utf8]{inputenc}
\usepackage{placeins}
\usepackage{algorithm}
\usepackage{algorithmic}

\usepackage{microtype}

\usepackage{inconsolata}

\usepackage{graphicx}
\usepackage{booktabs}
\usepackage{amsmath}
\usepackage{amssymb}

\title{Agentic Active Omni-Modal Perception for Multi-Hop Audio-Visual Reasoning}

\author{
    Ke Xu\textsuperscript{*},
    Yuhao Wang\textsuperscript{*},
    Ziyang Cheng,
    Hongcheng Liu \\
    {\bf Yanfeng Wang,
    Yu Wang\textsuperscript{\dag}} \\
    Shanghai Jiao Tong University \\
    \texttt{\{overji1, colane, muye12, hongcheng\_liu\}@sjtu.edu.cn}\\
    \texttt{\{wangyanfeng,yuwangsjtu\}@sjtu.edu.cn} 
}

\begin{document}
\maketitle
\begingroup
\renewcommand{\thefootnote}{\fnsymbol{footnote}}
\footnotetext[1]{Equal contributions.}
\footnotetext[2]{Corresponding authors.}
\endgroup
\begin{abstract}
Multi-hop audio-visual reasoning remains challenging for Omni-LLMs, as relevant evidence is often sparse, temporally dispersed, and distributed across both audio and visual streams. Existing benchmarks provide limited investigation of this setting, typically involving only a limited number of modalities, relevant temporal segments, or reasoning steps. In this work, we introduce MOV-Bench, a benchmark containing 519 carefully curated questions that require multi-hop reasoning over temporally dispersed audio-visual evidence. Evaluations on MOV-Bench reveal that current Omni-LLMs still struggle with multi-hop cross-modal reasoning. To address this challenge, we further propose AOP-Agent, an efficient agentic framework built on open-source Omni-LLMs for active omni-modal perception. By combining a hierarchical omni-modal memory with a collaborative observe-reflect-replan loop, AOP-Agent enables open-source Omni-LLMs to perform active perception without additional training or proprietary models. Experiments on MOV-Bench and OmniVideoBench demonstrate that AOP-Agent consistently improves reasoning performance, with particularly notable gains on long videos and reasoning-intensive questions.
\end{abstract}

\input{sections/Introduction_1}

\input{sections/Benchmark_2}
\input{sections/Method_3}

\input{sections/Experiment_4}

\input{sections/Conclusion_5}
\input{sections/Limitations_6}
\input{sections/Ethical_Considerations_7}

\bibliography{custom}

\clearpage
\appendix

\input{sections/Appendix/Related_Work}
\input{sections/Appendix/Analysis_of_MOV-Bench}
\input{sections/Appendix/Implementation}

\input{sections/Appendix/Details_of_AOP-Agent}
\input{sections/Appendix/CaseStudy}
\input{sections/Appendix/Prompts}

\end{document}

%% file: sections/Introduction_1.tex
\section{Introduction}

Recent end-to-end omni-modal large language models (Omni-LLMs)~\cite{zhao2025humanomni, li2025baichuan, Mingomni2025, Qwen3-Omni,qwenteam2026qwen35omnitechnicalreport,yang2025proagentharnessingondemandsensory,wang2025roboomni, xu2026reactiveproactiveassessingproactivity} have made substantial progress toward unified audio, visual, and textual understanding. Despite these advances, multi-hop audio-visual reasoning remains particularly challenging, since task-relevant evidence is often sparse, temporally dispersed, and distributed across both audio and visual streams. Answering such questions therefore requires cross-modal multi-hop reasoning over evidence appearing at different temporal locations.

\begin{figure}[h]
\centering
\includegraphics[width=0.9\linewidth]{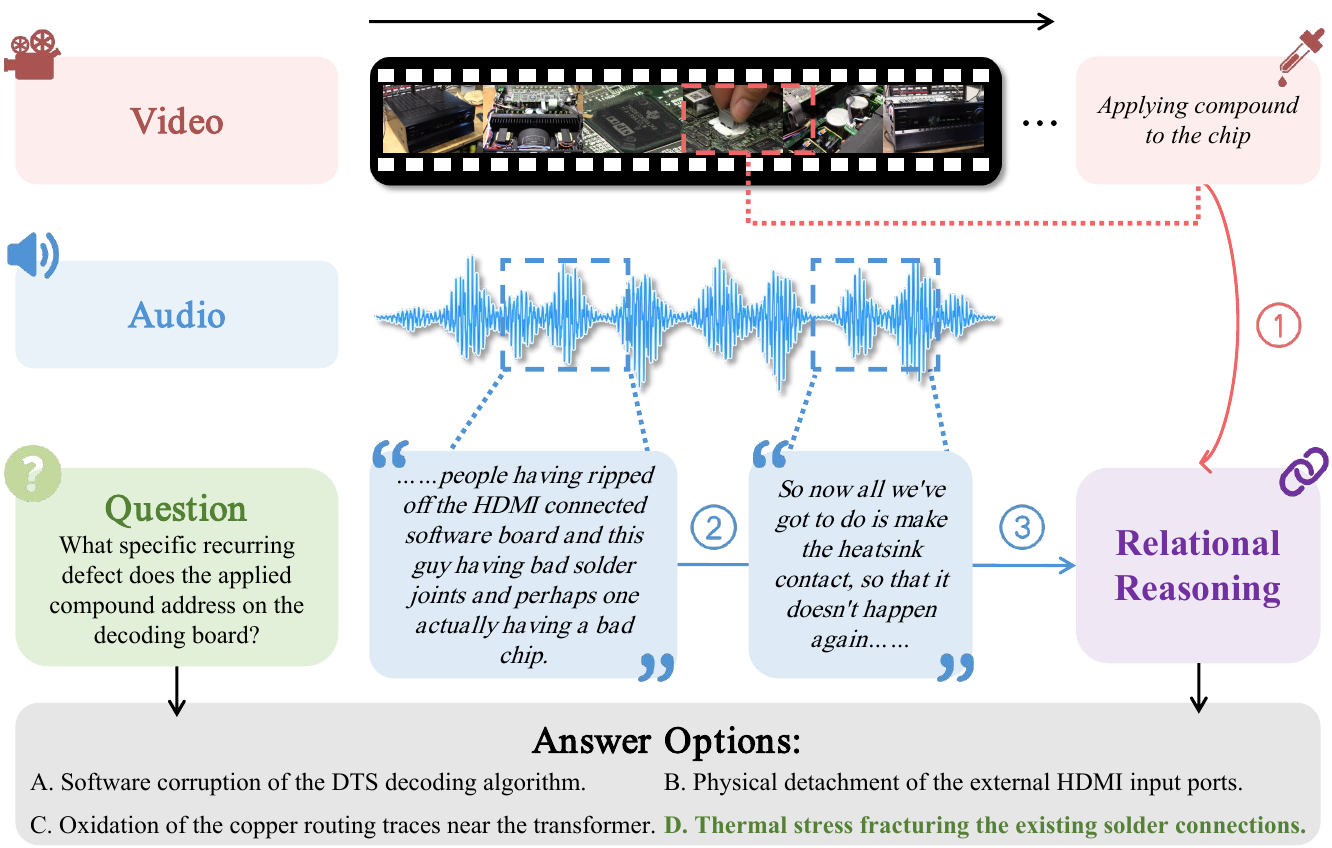}
\caption{A representative example from MOV-Bench illustrating audio-visual multi-hop reasoning in the video. The question requires linking temporally separated visual evidence, which identifies the compound applied to the chip, with audio evidence describing the bad chip and the heatsink contact. The correct answer can only be obtained by jointly aligning these sparse audio-visual cues across time.}
\label{fig:benchmark-example}
\end{figure}

Existing video benchmarks have begun to include audio-visual understanding tasks~\cite{li2025omnivideobenchaudiovisualunderstandingevaluation, hong2026worldsenseevaluatingrealworldomnimodal, wang2025proactivevideoqacomprehensivebenchmarkevaluating}. However, their task composition is often broad and heterogeneous, making it difficult to isolate the specific capability of cross-modal multi-hop reasoning in videos. Even when existing questions involve multi-hop reasoning over videos, the number of reasoning steps or relevant temporal segments is often limited, and the decisive evidence may still be concentrated in a single modality~\cite{chen2024cgbenchcluegroundedquestionanswering, fu2024video, fu2026video}. As a result, existing evaluations provide limited insight into whether Omni-LLMs can jointly integrate audio and visual evidence distributed across multiple temporal segments to answer complex multi-hop questions.

To address this gap, we propose \textbf{MOV-Bench} (\textbf{M}ulti-hop \textbf{O}mni \textbf{V}ideo Benchmark), a benchmark for multi-hop audio-visual reasoning across multiple temporal segments(Figure~\ref{fig:benchmark-example}). MOV-Bench contains 519 meticulously curated multiple-choice questions requiring models to integrate temporally dispersed evidence across modalities and video segments. Based on MOV-Bench, we find that the key challenge for current Omni-LLMs lies not only in reasoning over audio-visual evidence, but also in effectively locating and acquiring the sparse, temporally dispersed evidence required for complex multi-hop reasoning.


This observation motivates a shift from passive video processing toward active and iterative perception. Unlike existing Omni-LLMs that passively encode entire videos, human viewers often selectively attend to relevant moments, revisit earlier evidence, and progressively refine their understanding. Such active observation is particularly important when decisive evidence is sparse or temporally dispersed~\cite{wang2024lvbench, nguyen-etal-2024-video, zhang2026silvrsimplelanguagebasedvideo, qiu2026longvideor1smartnavigationlowcost}.

Recent agentic approaches have demonstrated the potential of active perception for audio-visual reasoning~\cite{luo2025videoragvisuallyalignedretrievalaugmentedlong, liu2025longvideoagentmultiagentreasoninglong, chen2026thinkgroundingcurriculumreinforced}. However, existing methods are often resource-intensive, either depending on costly proprietary models for multi-round decision making or requiring extensive training to equip open-source models with strong planning and reflection capabilities. This raises a key question: can open-source Omni-LLMs effectively achieve active perception without costly proprietary models or large-scale training?

\begin{figure}[t]
\centering
\includegraphics[width=0.87\linewidth]{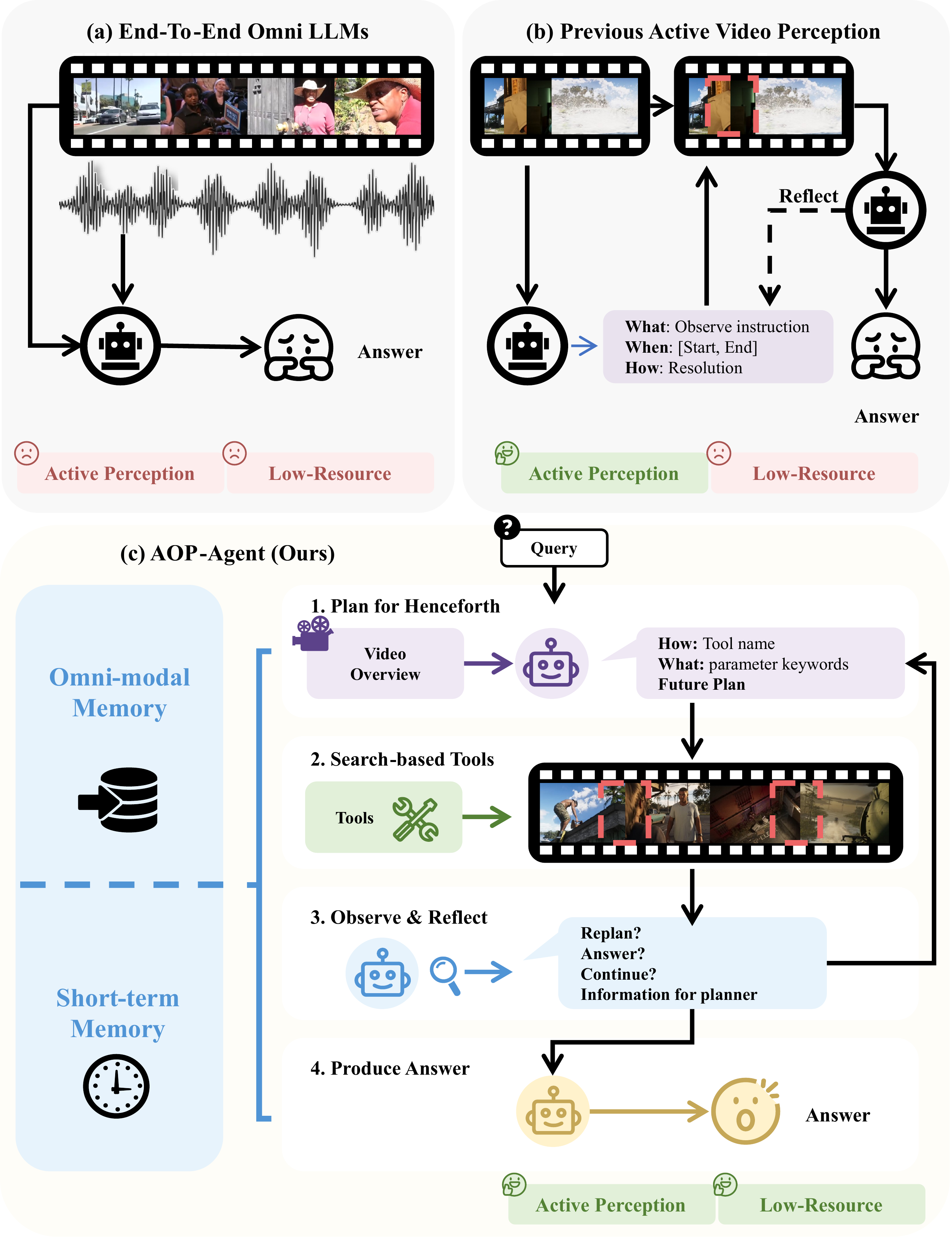}
\caption{\textbf{Motivation of AOP-Agent.} Existing methods either passively process the full video or depend on high-resource settings such as extensive training or proprietary models. AOP-Agent enables Omni-LLMs to actively decide which video evidence to observe under low-resource settings. Short-term Memory denotes working memory and evidence memory in AOP-Agent.}
\label{fig:motivation}
\end{figure}

To address this challenge, we propose \textbf{AOP-Agent} (\textbf{A}ctive \textbf{O}mni \textbf{P}erception Agent), an agentic framework for omni-modal reasoning under low-resource settings~(Figure~\ref{fig:motivation}). AOP-Agent reduces the difficulty of active perception for open-source Omni-LLMs by organizing videos into a hierarchical omni-modal memory and decomposing complex audio-visual reasoning into collaborative observe-reflect-replan interactions. This design enables open-source Omni-LLMs to progressively localize and integrate sparse audio-visual evidence without additional training or proprietary models.

Experiments demonstrate that AOP-Agent consistently improves multi-hop omni-modal reasoning, especially on long videos and reasoning-intensive questions. AOP-Agent achieves substantial gains on both MOV-Bench and  OmniVideoBench benchmark~\cite{li2025omnivideobenchaudiovisualunderstandingevaluation}. Further comparisons with representative agentic video reasoning frameworks show that AOP-Agent enables more efficient active perception for open-source Omni-LLMs.

The main contributions of this work are summarized as follows:

\begin{itemize}
\item We introduce \textbf{MOV-Bench}, a benchmark for multi-hop audio-visual reasoning that requires integrating temporally dispersed evidence across multiple video segments and modalities.
\item We propose \textbf{AOP-Agent}, a low-resource agentic framework that enables active perception for open-source Omni-LLMs through hierarchical omni-modal memory and collaborative observe-reflect-replan interactions, without additional training or proprietary models.
\item Experiments on MOV-Bench and OmniVideoBench demonstrate that AOP-Agent consistently improves multi-hop omni-modal reasoning, with particularly notable gains on long videos and reasoning-intensive questions.
\end{itemize}

%% file: sections/Benchmark_2.tex
\section{MOV-Bench}

\begin{figure*}[t]
  \centering
  \includegraphics[width=0.87\linewidth]{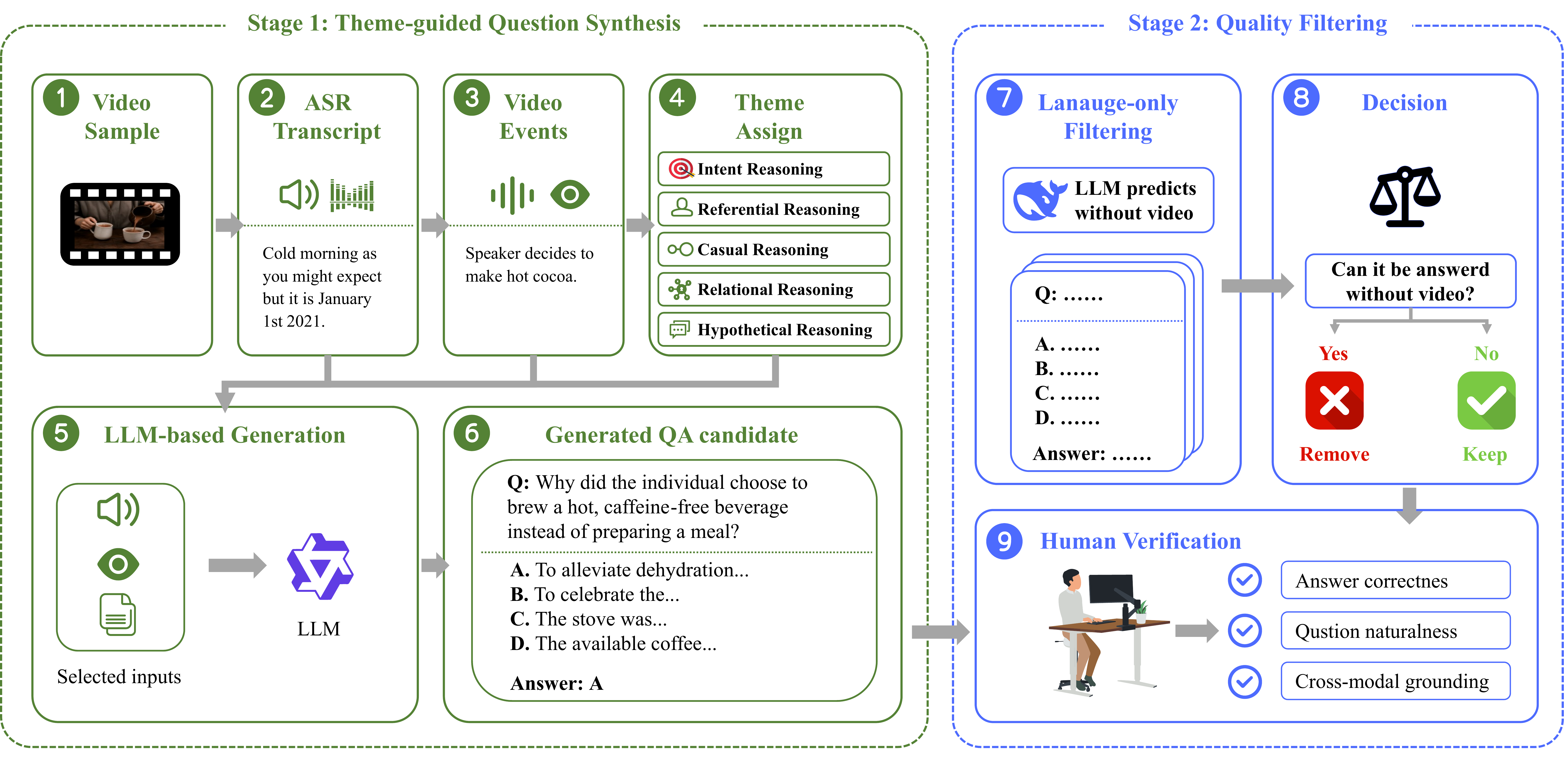}
  \caption{
  Overall pipeline for data collection, question synthesis, and quality filtering in MOV-Bench.
  }
  \label{fig:benchmark_pipeline}
\end{figure*}

MOV-Bench is designed to evaluate whether Omni-LLMs can solve cross-modal multi-hop questions by locating and integrating temporally dispersed audio-visual evidence. Unlike conventional video QA tasks that can often be answered from a single clip or a single modality, MOV-Bench requires models to connect multiple pieces of evidence across different video segments and modalities. As shown in Figure~\ref{fig:benchmark_pipeline}, our construction pipeline consists of theme-guided question synthesis followed by automatic and human quality filtering.

\subsection{Benchmark Design}

MOV-Bench is built around three design principles.

\noindent
\textbf{Multi-hop temporal reasoning.}
Each question requires evidence from multiple video segments rather than a single moment. The reasoning chains involve 2, 3, or 4 hops, allowing MOV-Bench to evaluate models beyond short-range perception.

\noindent
\textbf{Cross-modal evidence integration.}
Questions are constructed such that neither the visual stream nor the audio stream alone is sufficient. Models must jointly interpret visual events, speech content, and audio cues to infer the answer.

\noindent
\textbf{Reasoning-oriented question themes.}
To cover diverse reasoning patterns, questions are organized around five themes: causal, referential, relational, hypothetical, and intent reasoning. These themes define distinct reasoning patterns and help ensure that MOV-Bench covers diverse forms of audio-visual inference.

\subsection{Dataset Construction}

We construct MOV-Bench from Fine-Video~\cite{Farré2024FineVideo}, which contains diverse videos with rich visual and auditory content. As shown in Figure~\ref{fig:benchmark_pipeline}, we adopt an LLM-based question generation pipeline. For each video, we first provide the LLM with ASR transcripts, video event descriptions, and a randomly selected reasoning theme. The LLM is then prompted to generate multiple-choice questions whose answers require evidence from both modalities and multiple temporal segments. To avoid unnatural or weakly grounded questions, the LLM is instructed to return empty content when the video does not contain sufficient evidence for a valid cross-modal multi-hop question.

\subsubsection{Theme-guided Question Synthesis}

We define five reasoning themes to guide question generation.

\noindent
\textbf{Causal Reasoning} requires inferring the cause or consequence of an event by connecting audio and visual evidence.

\noindent
\textbf{Referential Reasoning} requires resolving references in one modality using evidence from another, such as identifying a visual entity mentioned or implied in speech.

\noindent
\textbf{Relational Reasoning} requires deriving social, spatial, or logical relationships between entities across modalities.

\noindent
\textbf{Hypothetical Reasoning} requires inferring possible outcomes under altered conditions based on facts established in the video.

\noindent
\textbf{Intent Reasoning} requires understanding a person's goal or motivation by jointly considering actions and spoken content.

\subsubsection{Quality Filtering}

We apply a two-stage filtering process to improve reliability. First, we perform a language-only filtering step: generated questions are given to LLMs without video input, and questions that can be answered from linguistic bias or prior knowledge are removed. Second, human annotators verify the remaining samples for answer correctness, question naturalness, and consistency between the question, options, and supporting video evidence. This process ensures that MOV-Bench emphasizes genuine cross-modal multi-hop reasoning rather than shortcut-based answering. The statistics of MOV-Bench is in Appendix \ref{sec:analysis_of_mov_bench}.

%% file: sections/Method_3.tex
\section{AOP-Agent}

MOV-Bench introduces a challenging setting for audio-visual reasoning, where answering questions requires integrating sparse audio-visual clues distributed across multiple temporal segments. A natural solution is active perception, where models iteratively identify and inspect relevant video segments during reasoning. However, existing approaches often rely on costly proprietary models or additional training to support such iterative observation.



Hence, we propose \textbf{AOP-Agent}, an efficient agentic framework for active omni-modal perception under low-resource settings. AOP-Agent combines hierarchical omni-modal memory construction with a multi-agent observe-reflect-replan loop, enabling open-source Omni-LLMs to progressively refine their understanding through iterative audio-visual observation without additional training or proprietary components. Detailed cases are provided in Appendix~\ref{sec:case_study}.

\subsection{Overview of AOP-Agent}
\begin{figure*}[t]
  \centering
  \includegraphics[width=0.89\linewidth]{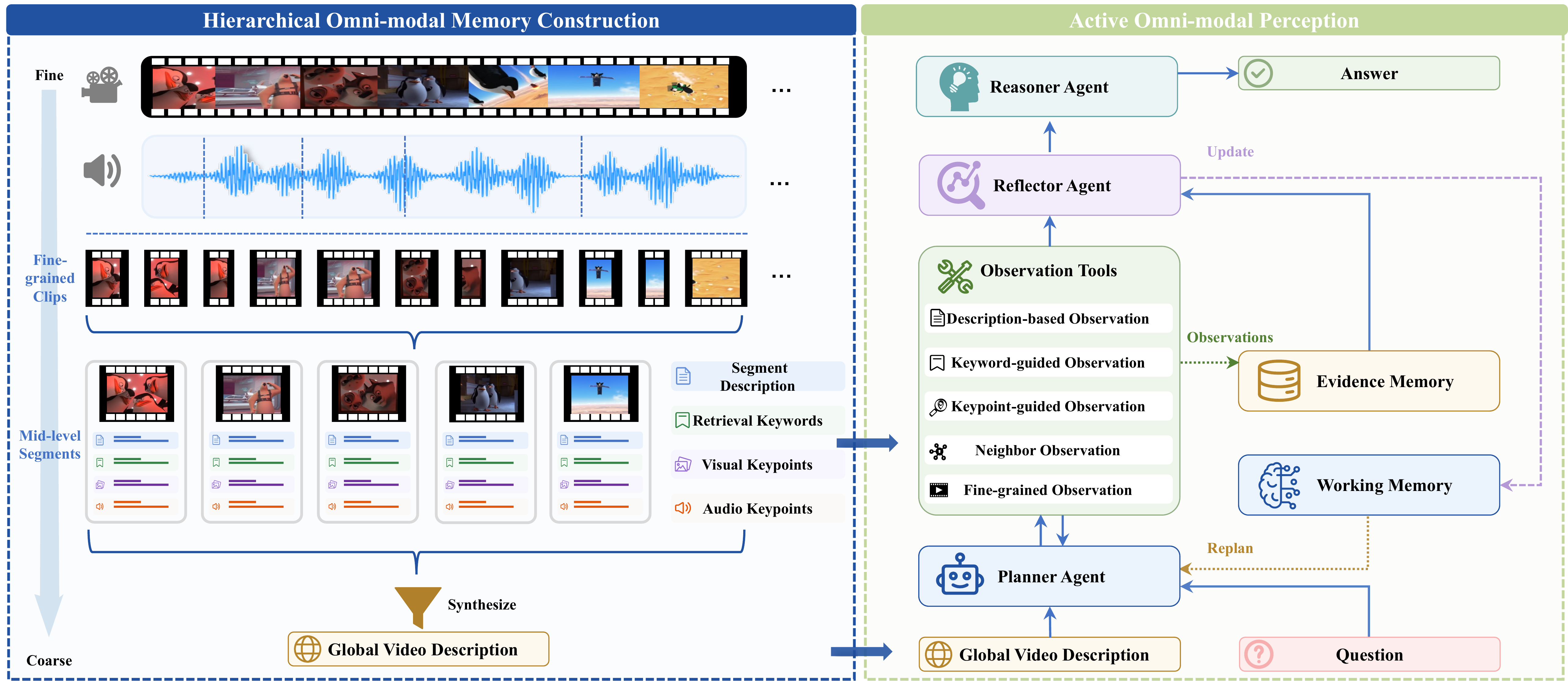}
  \caption{
  \textbf{Framework of AOP-Agent.}
  AOP-Agent reformulates omni-modal reasoning as active perception through hierarchical omni-modal memory and an observe-reflect-replan loop.
  }
  \label{fig:agent_framework}
\end{figure*}

As illustrated in Figure~\ref{fig:agent_framework}, AOP-Agent performs audio-visual reasoning through a multi-agent observe-reflect-replan process over hierarchical omni-modal memory. Given a long video $V$ and a natural language question $Q$, AOP-Agent first converts the raw video stream into a structured omni-modal memory $\mathcal{M}_V$. The agents then progressively observe, reflect, and replan over the memory to refine their understanding of the video before generating the final answer $A$.

\noindent
\textbf{Hierarchical Omni-modal Memory.}
AOP-Agent organizes the video into a hierarchical memory with multiple temporal and semantic granularities, including global video summaries, segment-level descriptions, audio-visual keypoints, and retrieval-oriented keywords. This hierarchical structure supports efficient localization and coarse-to-fine observation.

\noindent
\textbf{Active Omni-modal Perception.}
Based on the hierarchical memory $\mathcal{M}_V$, AOP-Agent performs iterative audio-visual perception through a multi-agent observe-reflect-replan loop. Specifically, the planner agent $A_P$ determines the next observation target according to the current reasoning state, specifying the video regions that should be further inspected. The observation tools $W_T$ then follow the planner's instruction to retrieve the corresponding mid-level segments. Based on these returned segments, the reflector agent $A_R$ evaluates whether the current information is sufficient or further observation is needed. Once sufficient information has been accumulated across iterations, the reasoner agent $A_A$ generates the final answer. More details are provided in Appendix~\ref{sec:details_agentic_framework}.

\subsection{Hierarchical Omni-modal Memory Construction}

Audio-visual reasoning in long videos is limited by context capacity and attention dilution.
Instead of directly processing the entire video stream, AOP-Agent organizes the video into a hierarchical omni-modal memory that supports coarse-to-fine evidence localization and observation.

Specifically, we apply a video segmentation strategy based on the timestamps of ASR results. 
The original video is split into $N_{\mathrm{fine}}$ fine-grained clips to preserve local audio-visual details as:
$
\{v^{\mathrm{fine}}_{i}
=
(b^{\mathrm{fine}}_{i}, e^{\mathrm{fine}}_{i})\}_{i=1}^{N_{\mathrm{fine}}},
$
where $b^{\mathrm{fine}}_{i}$ and $e^{\mathrm{fine}}_{i}$ denote the start and end timestamps of the $i$-th fine-grained clip $v_{i}$.
Adjacent clips are then merged into $N_{\mathrm{mid}}$ mid-level semantic segments, each no longer than 30 seconds, and compressed at a lower resolution as:
$
\{v^{\mathrm{mid}}_{i}
=
(b^{\mathrm{mid}}_{i}, e^{\mathrm{mid}}_{i})\}_{i=1}^{N_{\mathrm{mid}}}.
$
For each mid-level segment $v_i^{\mathrm{mid}}$, we use an Omni-LLM to generate structured semantic information, including visual keypoints $K_i^v$ that characterize events depicted in the visual content of the segment, audio keypoints $K_i^a$ that capture events present in the audio stream of the segment, retrieval keywords $K_i^w$ that concisely summarize the segment, and a segment-level description $D_{V,i}$ that provides a sentence-level summary of the segment.
The global video description $D_V$ is then synthesized from all segment-level descriptions:
$
D_V = \mathrm{LLM}\left(\{D_{V,i}\}_{i=1}^{N_{\mathrm{mid}}}\right).
$
Examples of the structured semantic information are provided in Appendix~\ref{sec:case_of_memory}.

The hierarchical memory provides both global video context and fine-grained audio-visual observations, enabling AOP-Agent to adaptively select observation granularity, and efficiently inspect relevant video content during iterative perception. More implementation details of the video segmentation strategy are provided in Appendix~\ref{sec:video_segmentation_imp}.

\subsection{Active Omni-modal Perception}

After constructing the hierarchical omni-modal memory, AOP-Agent performs active perception through a multi-agent observe-reflect-replan loop. Instead of uniformly processing the entire video at fine granularity, AOP-Agent dynamically determines where to observe and how finely to inspect according to the current reasoning state. To support flexible perception over long videos, AOP-Agent employs a set of complementary observation tools.

\noindent
\textbf{Description-based Observation Tool.}
This tool locates semantically relevant mid-level segments by matching planner queries against segment-level descriptions $\{D_{V,i}\}_{i=1}^{N_{\mathrm{mid}}}$. It is particularly effective for observing high-level events, intentions, and relational information.

\noindent
\textbf{Keyword-guided Observation Tool.}
This tool identifies segments containing explicit entities, actions, or spoken content through keyword-based matching over semantic keywords $\{K_i^w\}_{i=1}^{N_{\mathrm{mid}}}$.

\noindent
\textbf{Keypoint-guided Observation Tool.}
This tool matches visual keypoints $\{K_i^v\}_{i=1}^{N_{\mathrm{mid}}}$ and audio keypoints $\{K_i^a\}_{i=1}^{N_{\mathrm{mid}}}$ to identify cross-modal clues related to the current reasoning state.

\noindent
\textbf{Neighbor Observation Tool.}
Given an observed segment, this tool expands perception to temporally adjacent segments, allowing AOP-Agent to recover related events that occur before or after the current observation target.

\noindent
\textbf{Fine-grained Observation Tool.}
Given a selected mid-level segment, this tool further inspects its corresponding fine-grained clips for detailed local verification when coarse-grained observations are insufficient.

Together, these observation tools provide complementary ways to navigate the hierarchical memory, enabling AOP-Agent to progressively localize relevant video regions and adaptively switch between coarse-grained and fine-grained perception during iterative reasoning.

During inference, the Planner Agent $A_P$ determines the next observation target according to the question $Q$, the global video description $D_V$, and the current Working Memory $M_W$. The selected observation tool then retrieves the corresponding mid-level segments for further inspection and stores them in the Evidence Memory $M_E$. The Reflector Agent $A_R$ evaluates whether the current observations are sufficient for reasoning and provides feedback for subsequent replanning if necessary. Once sufficient information has been accumulated across iterations, the Reasoner Agent $A_A$ generates the final answer based on the collected observations.

Through the observe-reflect-replan loop, AOP-Agent progressively narrows its observation scope and performs adaptive multi-grained perception for reliable multi-hop cross-modal reasoning over long videos. By combining hierarchical omni-modal memory with collaborative multi-agent decomposition, AOP-Agent enabling open-source Omni-LLMs to perform effective iterative video reasoning without costly proprietary models or additional training. Details of the observation tools and the complete agentic framework are provided in Appendix~\ref{sec:formulation_of_tools} and Appendix~\ref{sec:details_agentic_framework}, respectively.

%% file: sections/Experiment_4.tex
\section{Experiment}


In this section, we present experimental results on MOV-Bench and OmniVideoBench to evaluate the capabilities of current Omni-LLMs and the effectiveness of AOP-Agent for cross-modal multi-hop video reasoning. Specifically, we study the following research questions:

\noindent
\textbf{RQ1:} What limitations of current Omni-LLMs are exposed by MOV-Bench in cross-modal multi-hop video reasoning?

\noindent
\textbf{RQ2:} Can AOP-Agent improve cross-modal multi-hop reasoning across different Omni-LLMs and benchmarks?

\noindent
\textbf{RQ3:} How does AOP-Agent differ from existing agentic video reasoning frameworks under open-source Omni-LLM settings?

\noindent
\textbf{RQ4:} How do planning, reflection, model assignment, and observation rounds contribute to the performance of AOP-Agent?

\input{sections/experiment_data/MOV-Bench}

\input{sections/experiment_data/OmniVideoBench}

\subsection{Experimental Setup}

\subsubsection{Benchmarks}

We conduct experiments on both MOV-Bench and OmniVideoBench. MOV-Bench evaluates cross-modal multi-hop reasoning over temporally dispersed audio-visual evidence, while OmniVideoBench serves as a more general audio-visual understanding benchmark for evaluating the generalization ability of Omni-LLMs and AOP-Agent. Additional benchmark details are provided in Appendix~\ref{sec:eval_symbol_imp}.

\subsubsection{Evaluated Models and Baselines}

We evaluate representative Omni-LLMs including Qwen3-Omni-Instruct, Qwen3-Omni-Thinking, Qwen3-Omni-Captioner~\cite{Qwen3-Omni}, Qwen2.5-Omni-7B, Qwen2.5-Omni-3B~\cite{xu2025qwen25omnitechnicalreport}, Ming-Lite-Omni-1.5~\cite{Mingomni2025}, Baichuan-Omni-1.5~\cite{li2025baichuan}, and MiniCPM-o-4.5~\cite{cui2026minicpmo45realtimefullduplex}.

For agentic baselines, we evaluate OmniAgent~\cite{tao2026activeperceptionagentomnimodal} and ActiveVideoPerception~\cite{wang2025activevideoperceptioniterative}, which represent recent frameworks for active perception and agentic audio-visual reasoning. To ensure fair comparison under open-source settings, both frameworks are re-implemented using Qwen3-Omni-Instruct as the backbone model. Additional implementation details are provided in Appendix~\ref{sec:baseline_imp}. 

\subsubsection{Implementation of AOP-Agent}

To evaluate AOP-Agent across different Omni-LLMs, we implement AOP-Agent using Qwen3-Omni-Thinking, Qwen3-Omni-Instruct, and Qwen2.5-Omni-7B as backbone models, where all agents share the same backbone unless otherwise specified.

For hierarchical memory construction, we use Qwen3-Omni-Instruct to generate segment-level audio-visual descriptions, keypoints, and global video summaries. We adopt FunASR~\cite{gao2023funasr} for ASR-based temporal segmentation. 

\subsection{Results on MOV-Bench (RQ1)}
Table~\ref{tab:mov_bench_combined} presents the main results on MOV-Bench. Overall, current open-source Omni-LLMs achieve relatively low performance, suggesting that cross-modal multi-hop reasoning remains highly challenging. For example, Qwen3-Omni-Instruct achieves only 52.79\% overall accuracy under direct inference, highlighting the difficulty of cross-modal multi-hop reasoning over long videos.


\noindent
\textbf{Long videos substantially increase reasoning difficulty.}
Across different Omni-LLMs, performance consistently decreases as video length increases. For example, Qwen3-Omni-Instruct drops from 62.43\% on short videos to 45.50\% on long videos subset. Similar trends are observed across other models, suggesting that current Omni-LLMs struggle to preserve and retrieve sparse evidence over long temporal spans.


\noindent
\textbf{Higher-hop questions remain challenging for current Omni-LLMs.}
Most Omni-LLMs achieve lower performance on 4-hop questions compared with their overall accuracy. For example, Qwen3-Omni-Thinking drops from 55.49\% overall accuracy to 44.19\% on 4-hop questions, while Ming-Lite-Omni-1.5 decreases from 49.52\% to 41.86\%. These results suggest that current Omni-LLMs still struggle to connect sparse audio-visual clues distributed across multiple temporal segments.

\noindent
\textbf{MOV-Bench reveals a key bottleneck in current Omni-LLMs.}
Taken together, these results suggest that a significant limitation of current Omni-LLMs lies not only in reasoning itself, but also in effectively locating, preserving, and integrating sparse cross-modal evidence distributed across long videos.

\subsection{Results of AOP-Agent (RQ2)}

Table~\ref{tab:mov_bench_combined} shows that AOP-Agent consistently improves direct-inference baselines across different Omni-LLM backbones on MOV-Bench, including Qwen3-Omni-Instruct, Qwen3-Omni-Thinking, and Qwen2.5-Omni-7B. The gains are especially pronounced on long videos and reasoning-intensive questions. For example, with Qwen3-Omni-Instruct, AOP-Agent improves long-video accuracy from 45.50\% to 60.85\%. Table~\ref{tab:omnivideobench_result} further shows that AOP-Agent generalizes to OmniVideoBench, where it also improves performance on long-video and reasoning-related subsets.

These results suggest that active perception is particularly beneficial for long-video and reasoning-intensive scenarios, where decisive evidence is often sparse and temporally dispersed. Under direct full-video inference, Omni-LLMs must preserve and retrieve sparse question-relevant evidence from long videos filled with irrelevant context, making multi-hop reasoning highly challenging. In contrast, AOP-Agent progressively narrows the observation space through iterative evidence acquisition and reflection, allowing the model to focus on question-relevant segments and perform more reliable cross-modal evidence integration for multi-hop reasoning.

\subsection{Comparison with Other Agentic  Frameworks (RQ3)}

We further compare AOP-Agent with OmniAgent and ActiveVideoPerception using Qwen3-Omni-Instruct as the backbone model. 

As shown in Table~\ref{tab:mov_bench_combined}, existing agentic frameworks perform substantially worse than direct inference on MOV-Bench. For example, OmniAgent and ActiveVideoPerception achieve only 38.34\% and 35.27\% overall accuracy, both below direct Qwen3-Omni-Instruct inference. Similar trends are also observed on OmniVideoBench~(Table~\ref{tab:omnivideobench_result}).

These results suggest that existing agentic video reasoning frameworks do not naturally transfer to open-source Omni-LLMs. Prior methods are typically designed around powerful proprietary models with strong planning and long-horizon reasoning capabilities. As a result, directly introducing complex iterative observation may instead amplify error accumulation for weaker open-source models under sparse and temporally dispersed evidence.

In contrast, AOP-Agent consistently outperforms both direct inference and existing agentic frameworks across benchmarks. The key difference is that AOP-Agent explicitly reduces the difficulty of active perception for open-source Omni-LLMs. Its hierarchical omni-modal memory makes sparse evidence easier to localize and retrieve, while the multi-agent collaboration framework decomposes active perception into simpler subtasks. Together, these designs make active perception substantially more tractable for open-source Omni-LLMs.

\subsection{Ablation Study (RQ4)}
We conduct ablation experiments on the long-video subsets of MOV-Bench and OmniVideoBench, where sparse and temporally dispersed evidence makes active perception particularly important, to analyze the role of planning, reflection, model assignment, and observation rounds in AOP-Agent.


\input{sections/experiment_data/Ablation}

\subsubsection{Effect of AOP-Agent Components}

Table~\ref{tab:ablation} presents the contribution of different components in AOP-Agent on long-video subset from MOV-Bench and OmniVideoBench.

Adding the Planner consistently improves performance over direct reasoning, suggesting that actively selecting observation targets is beneficial for long-video omni-modal reasoning. Further introducing the Reflector leads to the strongest performance, indicating that iterative reflection helps stabilize active perception by verifying current observations and guiding subsequent replanning when necessary.

\input{sections/experiment_data/Model-Comparison}

\subsubsection{Effect of Model Assignment}

Table~\ref{tab:four_model_comparison} analyzes different model assignments within AOP-Agent. Stronger planners and reflectors consistently improve performance even when the final reasoner remains unchanged, while weaker planners and reflectors noticeably limit overall performance despite using a stronger reasoner.

These results suggest that the effectiveness of active perception is largely determined before final reasoning begins. High-quality planning and reflection lead to more reliable observation trajectories and evidence selection, ultimately determining whether the reasoner receives useful evidence for complex multi-hop reasoning.

\input{sections/experiment_data/Max-Turns}

\subsubsection{Effect of Observation Rounds}


Figure~\ref{fig:max_round_comparison} shows that increasing the maximum observation rounds from 1 to 3 improves performance, indicating that iterative observation helps AOP-Agent recover missed evidence and refine its evidence memory. However, further increasing the round limit leads to saturation or slight degradation, suggesting that excessive observation can introduce noisy or irrelevant segments. Overall, the results suggest that active observation should remain bounded: AOP-Agent performs best when it acquires sufficient evidence without accumulating excessive irrelevant context.


%% file: sections/experiment_data/MOV-Bench.tex
\begin{table*}[t]
\centering
\resizebox{0.95\textwidth}{!}{
\begin{tabular}{lccccc|ccc|ccc|c}
\toprule
Model
& Causal & Ref & Hypo & Rel & Intent
& Long & Medium & Short
& 2hop & 3hop & 4hop
& Overall \\

\midrule
\multicolumn{13}{l}{\textbf{Baseline (Direct Inference)}} \\
Qwen3-Omni-Instruct
& 48.35 & \textbf{59.63} & 35.00 & 54.33 & 55.00
& 45.50 & 49.65 & 62.43
& 54.92 & 49.28 & 51.16
& 52.79 \\

Qwen3-Omni-Thinking
& 56.04 & 57.14 & 35.00 & 60.63 & 58.75
& 48.15 & 51.77 & 65.61
& 57.29 & 58.70 & 44.19
& 55.49 \\

Qwen3-Omni-Captioner
& 59.34 & 57.76 & 30.00 & 51.97 & 57.50
& 46.56 & 53.19 & 60.32
& 52.88 & 54.35 & 53.49
& 53.37 \\

Qwen2.5-Omni-7B
& 41.76 & 52.80 & 25.00 & 55.12 & 42.50
& 37.57 & 45.39 & 56.61
& 49.15 & 45.65 & 39.53
& 46.63 \\

Qwen2.5-Omni-3B
& 42.86 & 45.96 & 38.33 & 44.09 & 43.75
& 35.98 & 43.97 & 51.32
& 44.41 & 46.38 & 37.21
& 43.74 \\

Ming-Lite-Omni-1.5
& 46.15 & 55.90 & 35.00 & 50.39 & 50.00
& 44.44 & 47.52 & 56.08
& 49.83 & 53.62 & 41.86
& 49.52 \\

Baichuan-Omni-1.5
& 41.76 & 47.83 & 30.00 & 44.09 & 47.50
& 37.57 & 39.72 & 52.91
& 45.76 & 45.65 & 33.72
& 43.74 \\

MiniCPM-o 4.5
& 54.95 & 58.39 & 40.00 & 54.33 & 48.75
& 43.39 & 53.90 & 62.43
& 56.27 & 50.00 & 47.67
& 53.18 \\


\midrule
\multicolumn{13}{l}{\textbf{Other Agentic Framework}} \\
OmniAgent
& 45.05 & 38.51 & 28.33 & 35.43 & 42.50
& 33.86 & 39.72 & 41.80
& 36.61 & 45.65 & 32.56
& 38.34 \\

Active Video Perception
& 38.46 & 32.30 & 34.48 & 30.16 & 46.25
& 32.80 & 38.85 & 35.11
& 33.79 & 39.42 & 33.72
& 35.27 \\



\midrule
\multicolumn{13}{l}{\textbf{AOP-Agent (Ours)}} \\
Qwen3-Omni-Instruct
& \textbf{71.43} & 58.39 & \textbf{50.00} & 62.99 & \textbf{70.00}
& \textbf{60.85} & 62.41 & 64.55
& \textbf{64.75} & 61.59 & 56.98
& \textbf{62.62} \\

Qwen3-Omni-Thinking
& 70.33 & 55.28 & 48.33 & \textbf{68.50} & 67.50
& 55.03 & \textbf{63.12} & \textbf{68.78}
& 60.68 & \textbf{67.39} & \textbf{59.30}
& 62.24 \\

Qwen2.5-Omni-7B
& 50.55 & 47.83 & 41.67 & 51.18 & 38.75
& 41.27 & 46.81 & 52.91
& 48.14 & 47.10 & 43.02
& 47.01 \\
\bottomrule
\end{tabular}
}
\caption{
Results of different models and frameworks. The table reports accuracy on videos across five reasoning types, three duration ranges and three hop types. Boldface highlights the best performace within each column.
}
\label{tab:mov_bench_combined}
\end{table*}

%% file: sections/experiment_data/OmniVideoBench.tex
\begin{table}[t]
\centering
\resizebox{0.95\linewidth}{!}{
\begin{tabular}{lccc|c|c}
\toprule
Model
& Long & Medium & Short & Reason & Overall \\

\midrule
\multicolumn{6}{l}{\textbf{Baseline (Direct Inference)}} \\
Qwen3-Omni-Instruct
& 28.45 & 26.27 & 38.21 & 31.09 & 29.30\\
Qwen3-Omni-Thinking
& 31.90 & \textbf{37.29} & 37.40 & 35.57 & 31.90\\
Qwen2.5-Omni-7B
& 30.17 & 27.12 & 34.15 & 30.53 & 29.00\\

\midrule
\multicolumn{6}{l}{\textbf{Other Video Agent Frameworks}} \\
OmniAgent
& 28.45 & 34.75 & 31.71 & 31.65 & 27.30\\
Active Video Perception
& 30.17 & 22.88 & 36.89 & 30.06 & 26.90\\
\midrule

\multicolumn{6}{l}{\textbf{AOP-Agent (Ours)}} \\
 Qwen3-Omni-Instruct
& \textbf{40.52} & 33.05 & \textbf{39.84} & 37.82 & \textbf{33.20}\\
 Qwen3-Omni-Thinking
& \textbf{40.52} & 35.59 & \textbf{39.84} & \textbf{38.66} & 32.90\\
 Qwen2.5-Omni-7B
& 31.90 & 33.05 & 34.15 & 33.05 & 31.10\\
\bottomrule
\end{tabular}
}
\caption{
Comparison of the baselines and AOP-Agent on OmniVideoBench. Boldface highlights the best performace within each column.
}
\label{tab:omnivideobench_result}
\end{table}

%% file: sections/experiment_data/Ablation.tex
\begin{table}[t]
\centering
\resizebox{\linewidth}{!}{
\begin{tabular}{lcccc}
\toprule
Component          & MOV Long & OVB Long \\
\midrule
Reasoner (Baseline)              & 45.50  & 28.45 \\
+ Planner                      & 50.26 & 34.48\\
+ Planner + Reflector (Full)  &  \textbf{60.85} & \textbf{40.52}\\
\bottomrule
\end{tabular}
}
\caption{Ablation Study on long-video subset of OmniVideoBench (OVB) and MOV-Bench (MOV).}
\label{tab:ablation}
\end{table}

%% file: sections/experiment_data/Model-Comparison.tex
\begin{table}[t]
\centering
\resizebox{\linewidth}{!}{
\begin{tabular}{llcc}
\toprule
Planner \& Reflector & Reasoner & MOV Long & OVB Long \\
\midrule
Qwen2.5-Omni-7B & Qwen2.5-Omni-7B & 41.27 & 31.90 \\
Qwen3-Omni-30B & Qwen2.5-Omni-7B & 48.68 & 33.62 \\
Qwen2.5-Omni-7B & Qwen3-Omni-30B & 58.20 & 35.34 \\
Qwen3-Omni-30B & Qwen3-Omni-30B & \textbf{60.85} & \textbf{40.52} \\
\bottomrule
\end{tabular}
}
\caption{Performance comparison of different model assignments on MOV-Bench (MOV) and OmniVideoBench (OVB).}
\label{tab:four_model_comparison}
\end{table}

%% file: sections/experiment_data/Max-Turns.tex

\begin{figure}[t]
\centering
\includegraphics[width=1.0\linewidth]{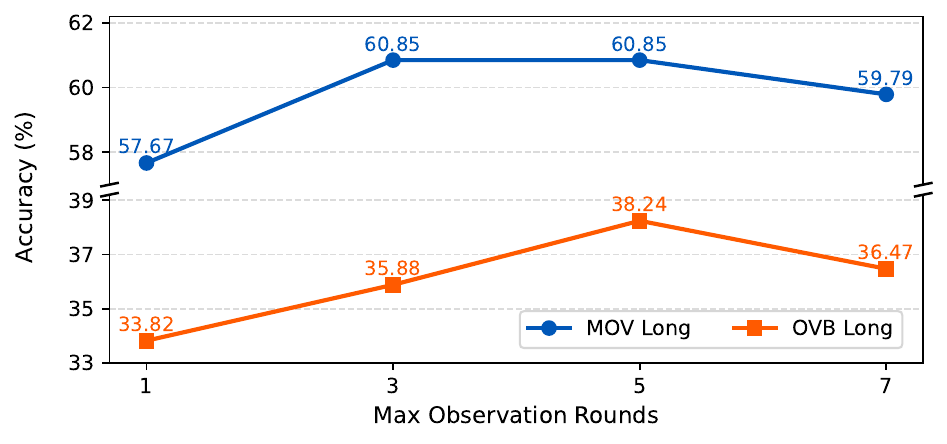}
\caption{Performance of AOP-Agent based on Qwen3-Omni-Instruct on long-video subsets under different maximum observation round limits.}
\label{fig:max_round_comparison}
\end{figure}

%% file: sections/Conclusion_5.tex
\section{Conclusion}

We introduce MOV-Bench, a benchmark for evaluating multi-hop audio-visual reasoning in videos, and AOP-Agent, an agentic framework for active omni-modal perception under low-resource settings. Results on MOV-Bench show that current Omni-LLMs are limited not only by reasoning over audio-visual evidence, but also by their ability to locate and acquire sparse evidence distributed across audio and vision streams. By combining hierarchical omni-modal memory with an observe-reflect-replan loop, AOP-Agent enables models to selectively inspect question-relevant evidence without additional training or proprietary models. Extensive experiments on MOV-Bench and OmniVideoBench demonstrate consistent improvements across multiple Omni-LLMs, highlighting active perception as a promising paradigm for scalable and effective long-video omni-modal reasoning.


%% file: sections/Limitations_6.tex
\section*{Limitations}
Although MOV-Bench and AOP-Agent advance omni-modal video understanding and reasoning, several limitations remain. 
First, due to the cost of annotation and verification, the scale of MOV-Bench remains limited. 
Although human verification is included in the construction pipeline, synthesized questions may still be less diverse and natural than fully human-labeled data. 
Second, although AOP-Agent improves performance on both MOV-Bench and OmniVideoBench, its multi-round observe-reflect-replan loop introduces additional inference overhead.
Third, as illustrated by the failure case in Figure~\ref{fig:failure_case}, hallucination and error propagation remain critical challenges in the omni-modal reasonings. 
Although the reflector can partially mitigate error propagation, the overall accuracy is still constrained by the quality of the hierarchical omni-modal memory base and the reliability of the underlying OmniLLM. 
Finally, the proposed omni-modal memory base is not streaming-friendly, as it relies on a multi-level offline construction process. 
This may limit its applicability to real-time scenarios such as smart glasses and live streaming.

%% file: sections/Ethical_Considerations_7.tex
\section*{Ethical Considerations}
MOV-Bench is constructed from Fine-Video~\cite{Farré2024FineVideo}, a publicly available dataset released for academic use. We comply with the license terms of Fine-Video, use the videos solely for research purposes, and do not redistribute any raw video content. Our benchmark construction process derives information from the original videos and does not intentionally introduce any new personally identifiable information.
Due to the scale of MOV-Bench, the authors served as human annotators. Human annotators were involved in verifying the quality, correctness, and evidence alignment of the generated samples. 
We used large language models as writing assistants to edit and polish the paper. The authors reviewed and revised the generated text and remain fully responsible for the content, claims, and integrity of the work. Overall, we believe that this work introduces minimal additional ethical or privacy risks beyond those already associated with the underlying Fine-Video dataset.

%% file: sections/Appendix/Related_Work.tex
\section{Related Work}

\textbf{Multimodal Benchmarks.}
Benchmarks for multimodal large language models have evolved from basic video perception~\cite{wang2024videoagentlongformvideounderstanding} to more comprehensive audio-visual understanding and agentic reasoning~\cite{wang2025activevideoperceptioniterative, tao2026activeperceptionagentomnimodal, zhu2026omniragagentagenticomnimodalreasoning}. 
Recent omni-modal benchmarks, including OmniVideoBench~\cite{li2025omnivideobenchaudiovisualunderstandingevaluation}, Daily-Omni~\cite{zhou2025dailyomni}, and WorldSense~\cite{hong2026worldsenseevaluatingrealworldomnimodal}, evaluate video understanding with both visual and auditory inputs. 
Other benchmarks, such as Video-MME~\cite{fu2024video}, FutureOmni~\cite{chen2026futureomnievaluatingfutureforecasting}, and Video-DR~\cite{liu2026watching}, further extend evaluation to comprehensive video understanding, future forecasting, and open-world video reasoning. 
OmniMMI~\cite{wang2025omnimmicomprehensivemultimodalinteraction} studies multimodal interaction, while OmniGAIA~\cite{li2026omnigaianativeomnimodalai} evaluates tool-integrated omni-modal agents. 
Different from these benchmarks, MOV-Bench focuses on multi-hop audio-visual reasoning, where models must integrate temporally dispersed evidence from both audio and visual streams.

\noindent
\textbf{Agentic Frameworks for Video Understanding.}
To mitigate long-context limitations, prior work has explored caption-based and retrieval-based video representations~\cite{fan2025agentickeyframesearchvideo, Wang_2025_CVPR, park2026framesusefulefficientstrategies}. 
These methods reduce input length by converting video segments into textual descriptions, but they usually rely on static representations and lack adaptive evidence acquisition. 
Recent agentic systems introduce planning, grounding, and reflection into video reasoning~\cite{chen2026thinkgroundingcurriculumreinforced}, including LVAgent~\cite{chen2025lvagentlongvideounderstanding}, ReAgent-V~\cite{zhou2025reagentvrewarddrivenmultiagentframework}, Active Video Perception~\cite{wang2025activevideoperceptioniterative}, and OmniAgent~\cite{tao2026activeperceptionagentomnimodal}. 
However, many of these methods depend on costly proprietary models or extensive training, making them resource-intensive.
AOP-Agent instead adopts an agentic active perception framework, enabling models to iteratively decide what evidence to observe through planning and reflection with low resource requirements.

%% file: sections/Appendix/Analysis_of_MOV-Bench.tex
\section{Statistics of MOV-Bench}
\label{sec:analysis_of_mov_bench}

\begin{figure*}[t]
  \centering
  \includegraphics[width=0.37\linewidth]{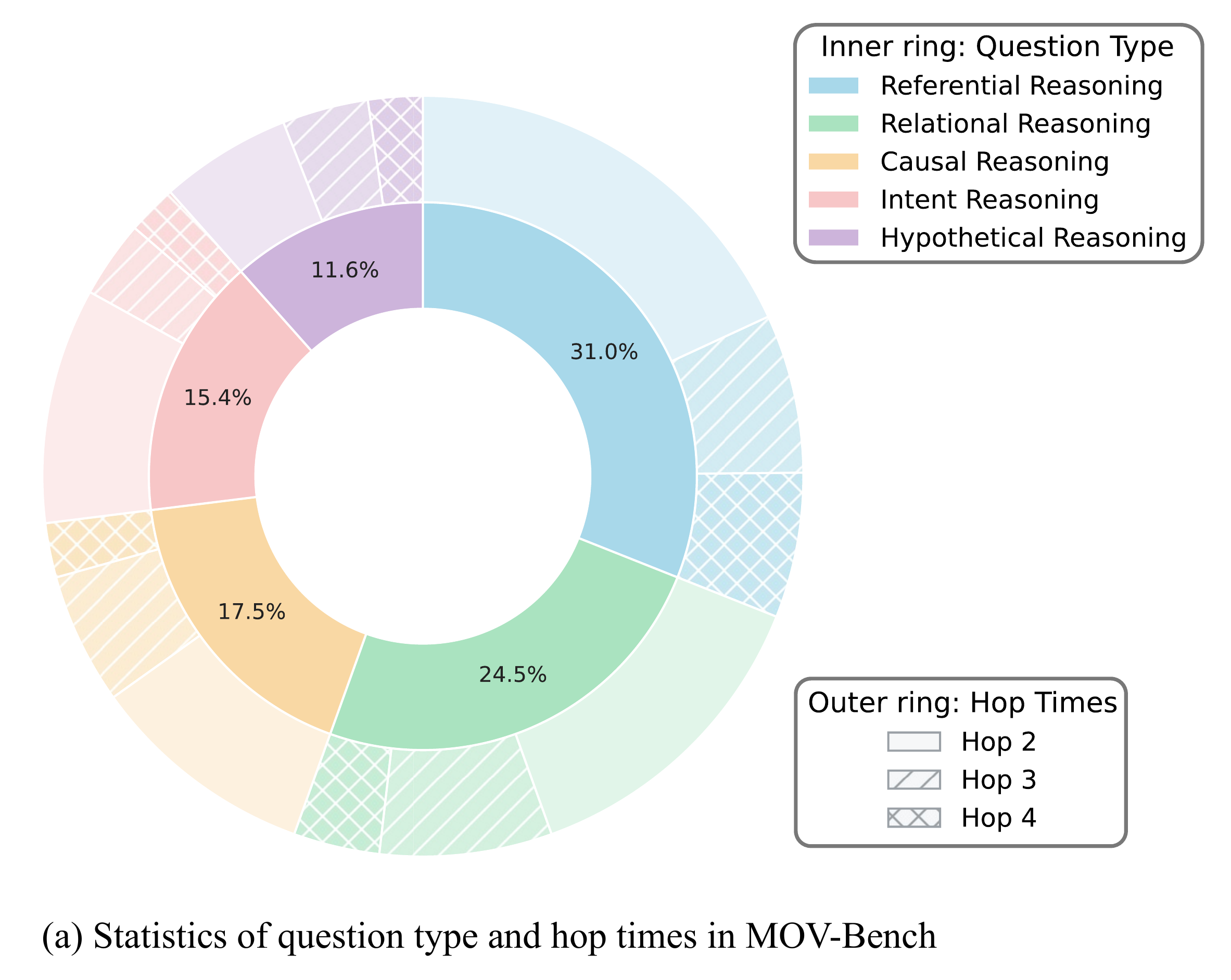}
  \includegraphics[width=0.6\linewidth]{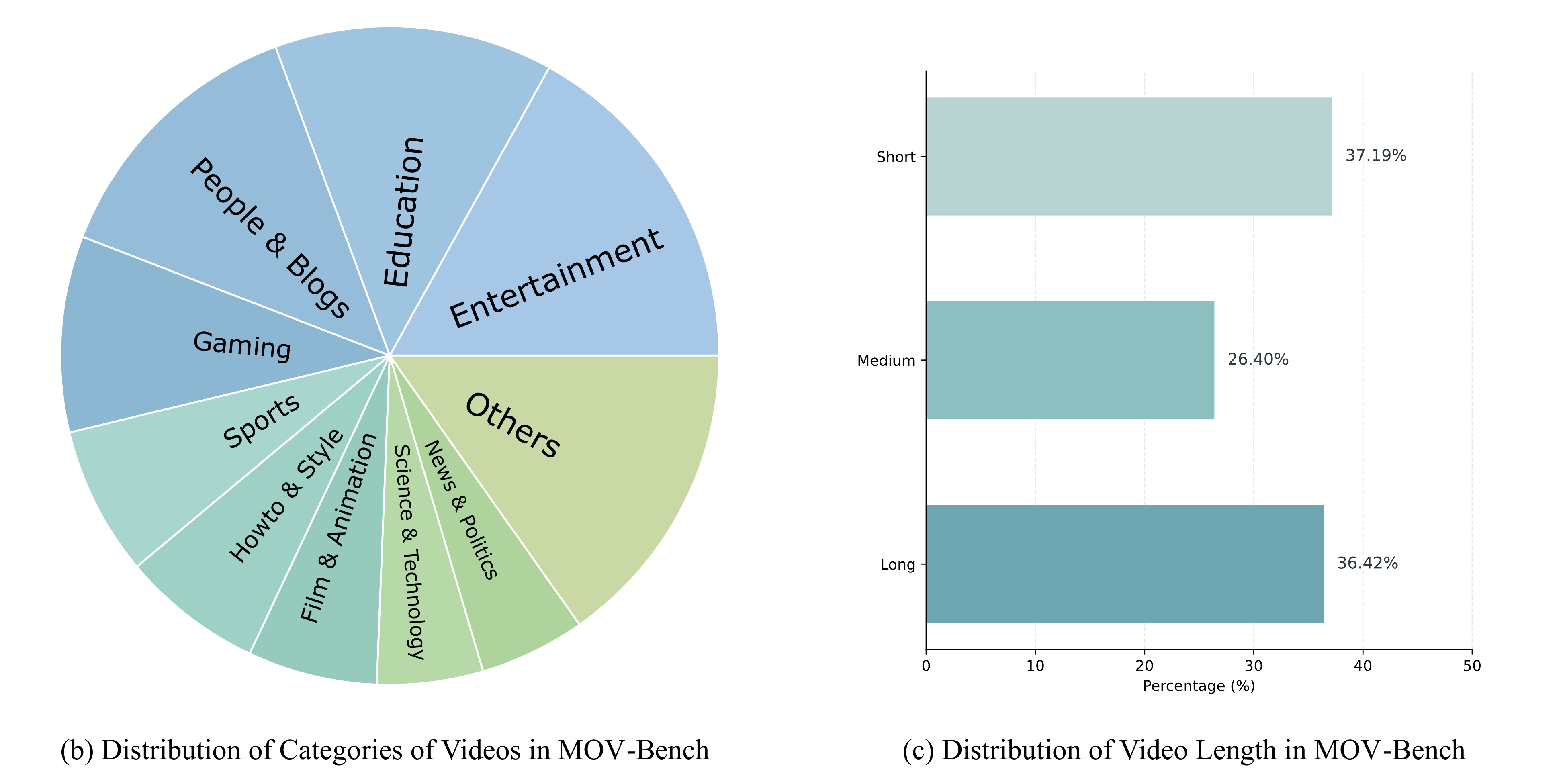}
  \caption{
  Statistics of MOV-Bench. 
  (a) MOV-Bench covers five reasoning-intensive question types with reasoning chains ranging from 2 to 4 hops. 
  (b) Videos span 15 real-world categories. 
  (c) Video duration distribution, where \textbf{Short} denotes videos shorter than 150\,s, \textbf{Medium} denotes videos of 150--300\,s, and \textbf{Long} denotes videos longer than 300\,s.
  }
  \label{fig:benchmark_statistics}
\end{figure*}

With videos from Fine-Video, their ASR-transcription and visual events within video events is provided in Fine-Video and audio events within video events is constructed with Qwen3-Omni-Instruct.
We initially generate 2,240 candidate samples using Qwen3.5-Plus~\cite{qwen35blog}. 
After preliminary multiple-choice validation with DeepSeek-V4-Flash~\cite{deepseekai2026deepseekv4}, 802 samples are retained. 
Following final human expert verification, MOV-Bench contains 519 high-quality multiple-choice questions.

As shown in Figure~\ref{fig:benchmark_statistics}, MOV-Bench covers 15 real-world video categories. 
The average video duration is 240.26 seconds, and long videos exceeding 5 minutes account for 36.42\% of the benchmark. 
The average question length is 17.02 words, while the average answer length is 7.87 words. 
To ensure multi-hop reasoning difficulty, MOV-Bench includes 295 two-hop questions, 138 three-hop questions, and 86 four-hop questions, with an average of 2.60 reasoning hops per sample.

%% file: sections/Appendix/Implementation.tex
\section{Experiment Details}
\label{sec:imp_details}

\subsection{Baselines}
\label{sec:baseline_imp}
AOP-Agent follows an agentic active perception paradigm. 
To compare AOP-Agent with representative agentic video understanding frameworks, we select OmniAgent and ActiveVideoPerception as baselines. 
For ActiveVideoPerception, we replace the original Gemini backbone with Qwen3-Omni-Instruct. 
For OmniAgent, whose audio and video tools may use different Omni-LLMs as backbone models, we replace the original OpenAI or Gemini models with Qwen3-Omni-Instruct. 
The brain LLM is also replaced with Qwen3-Omni-Instruct to ensure that the entire framework uses the same backbone. 
All other settings of OmniAgent and ActiveVideoPerception are kept unchanged.

\subsection{Hyperparameters}
\label{sec:hyperparameters_imp}
During hierarchical omni-modal memory construction, the LLM temperature is set to 1.0 to encourage diverse video descriptions. 
During the agentic observation process, the temperature is set to 0.2 to improve reproducibility. 
The maximum context length is set to 32,768 tokens to balance computational efficiency and long-context understanding.

\subsection{Evaluation Details}
\label{sec:eval_symbol_imp}
To further evaluate AOP-Agent on reasoning-related tasks, we conduct experiments on MOV-Bench and OmniVideoBench~\cite{li2025omnivideobenchaudiovisualunderstandingevaluation}, with the results reported in Table\ref{tab:omnivideobench_result}. 
Following the video-length categorization, videos longer than 300\,s are classified as \textit{Long}, videos between 150\,s and 300\,s as \textit{Medium}, and videos shorter than 150\,s as \textit{Short}. 
The \textit{Reason} subset denotes reasoning-related tasks, including causal reasoning, ego reasoning, hypothetical reasoning, relationship reasoning, and reference reasoning in OmniVideoBench. 
For the length-based subsets, i.e., \textit{Long}, \textit{Medium}, and \textit{Short}, we report only reasoning-related samples. 
The \textit{Overall} column reports performance on all samples, regardless of whether they belong to reasoning-related categories.

Additionally, in Table\ref{tab:mov_bench_combined}, “Casual,” “Ref,” “Hypo,” “Rel,” and “Intent” refer to causal, referential, hypothetical, relational, and intent reasoning, respectively. “2hop,” “3hop,” and “4hop” indicate the number of reasoning hops required by each sample in MOV-Bench.

%% file: sections/Appendix/Details_of_AOP-Agent.tex
\section{AOP-Agent Details}
\subsection{Cases of Hierarchical Omni-modal Memory}
\label{sec:case_of_memory}
For each mid-level segment $v_i^{\mathrm{mid}}$, we use an Omni-LLM to generate structured semantic information. We provide examples for each type of semantic information: a visual keypoint may be “The galaxy image shows swirling arms with bright star clusters and dark dust lanes”; an audio keypoint may be “A low, sustained, electronic musical tone plays throughout the segment”; retrieval keywords may include “galaxy,” “cosmos,” and “ambient music”; and the corresponding segment-level description may be “A static image of a spiral galaxy with the text ‘Spiral Arm Density Wave Theory’ overlaid, accompanied by ambient electronic music.”

\subsection{Tools in AOP-Agent}
\label{sec:formulation_of_tools}

Given a planner-generated query $q_t$ at interaction round $t$, the observation toolset $W_T$ maps the query and the current memory state to a set of candidate evidence segments:
\begin{equation}
\mathcal{O}_t = W_T(q_t, \mathcal{M}_V, M_E),
\end{equation}
where $\mathcal{M}_V$ denotes the hierarchical omni-modal memory and $M_E$ denotes the evidence memory accumulated in previous rounds.
The observation toolset contains five complementary operations.

\noindent
\textbf{Description-based Observation Tool.}
This tool retrieves mid-level segments whose segment-level descriptions are semantically relevant to the planner query.
Formally, for each mid-level segment $v_i^{\mathrm{mid}}$, we compute
\begin{equation}
s_{\mathrm{desc}}(q_t, i)
=
\cos \bigl( \phi(q_t), \phi(D_{V,i}) \bigr),
\end{equation}
where $\phi(\cdot)$ denotes the text embedding model BGE-M3~\cite{bge-m3} and $D_{V,i}$ is the segment-level description.
The tool returns the top-$k$ segments with the highest scores:
\begin{equation}
\mathcal{O}_{\mathrm{desc}}(q_t)
=
\operatorname{TopK}_{i \in [1,N_{\mathrm{mid}}]}
s_{\mathrm{desc}}(q_t, i).
\end{equation}
This tool is suitable for locating evidence described by high-level events, intentions, or relations.

\noindent
\textbf{Keyword-guided Observation Tool.}
This tool performs sparse lexical retrieval over the keyword memory.
Given the planner query $q_t$, AOP-Agent first derives a set of query keywords $K_t^q$ and then scores each segment by BM25 over its retrieval keywords $K_i^w$:
\begin{equation}
s_{\mathrm{kw}}(q_t, i)
=
\operatorname{BM25}(K_t^q, K_i^w).
\end{equation}
The returned observations are
\begin{equation}
\mathcal{O}_{\mathrm{kw}}(q_t)
=
\operatorname{TopK}_{i \in [1,N_{\mathrm{mid}}]}
s_{\mathrm{kw}}(q_t, i).
\end{equation}
This tool is useful when the question contains explicit entities, actions, or spoken terms.

\noindent
\textbf{Keypoint-guided Observation Tool.}
This tool searches both visual and audio keypoints to locate cross-modal evidence more directly.
For each segment, we compute dense semantic similarity and sparse lexical matching scores over visual and audio keypoints:
\begin{equation}
\begin{aligned}
s_{\mathrm{kp}}(q_t, i)
=
&\lambda
\max_{K \in \{K_i^v, K_i^a\}}
\cos \bigl( \phi(q_t), \phi(K) \bigr) \\
&+
(1-\lambda)
\max_{K \in \{K_i^v, K_i^a\}}
\operatorname{BM25}(q_t, K),
\end{aligned}
\end{equation}
where $\lambda \in [0,1]$ controls the balance between dense semantic matching and sparse lexical matching decided by planner $A_P$.
The tool returns
\begin{equation}
\mathcal{O}_{\mathrm{kp}}(q_t)
=
\operatorname{TopK}_{i \in [1,N_{\mathrm{mid}}]}
s_{\mathrm{kp}}(q_t, i).
\end{equation}
This operation is designed to retrieve segments containing visual or audio evidence directly related to the query.

\noindent
\textbf{Neighbor Observation Tool.}
Given an already observed segment $v_j^{\mathrm{mid}}$, this tool expands the observation to its temporal neighbors.
For a candidate neighboring segment $v_i^{\mathrm{mid}}$, the returned neighboring segments are
\begin{equation}
\mathcal{O}_{\mathrm{nbr}}(j)
=
\{v_i^{\mathrm{mid}} \mid 1 \le i \le N_{\mathrm{mid}},\ 0 < |i-j| \le r \},
\end{equation}
where $r$ is the neighbor radius decided by planner agent $A_P$ and $\mathbb{I}(\cdot)$ is the indicator function.
This tool helps recover temporally related evidence that may appear immediately before or after the initially retrieved segment.

\noindent
\textbf{Fine-grained Observation Tool.}
Given a selected mid-level segment $v_j^{\mathrm{mid}}$, this tool inspects the fine-grained clips contained within it.
The fine-grained observation tool returns the set of fine-grained clips whose timestamps fall inside the temporal span of $v_j^{\mathrm{mid}}$:
\begin{equation}
\mathcal{O}_{\mathrm{fine}}(j)
=
\{v_l^{\mathrm{fine}} \mid
b_j^{\mathrm{mid}} \le b_l^{\mathrm{fine}},
\ e_l^{\mathrm{fine}} \le e_j^{\mathrm{mid}}
\}.
\end{equation}

It is used when the model needs higher-resolution evidence for local verification.

\subsection{Agentic Framework of AOP-Agent}
\label{sec:details_agentic_framework}
AOP-Agent performs iterative video understanding through an observe-reflect-replan loop, where multiple agents collaborate to actively acquire question-relevant evidence.

\noindent \textbf{Working Memory.}
The working memory $M_W$ maintains the reasoning state across observation rounds, including historical planning trajectories, future plans, and reflection feedback:
\[
M_W = \{P_{\mathrm{past}}, P_{\mathrm{fut}}, R_r\},
\]
where $P_{\mathrm{past}}$ denotes historical planning states, $P_{\mathrm{fut}}$ denotes the future observation plan, and $R_r$ denotes the reflection feedback $R_r$ produced by the Reflector Agent $A_R$. Specifically, $R_r$ records why the current observations are insufficient and what should be inspected in the next replanning round. This memory enables the system to avoid redundant observations and maintain coherent reasoning progress.

\noindent \textbf{Evidence Memory.}
The evidence memory $M_E$ accumulates validated evidence segments collected during interaction. These segments are ranked according to their relevance scores and are updated at each round using observations returned by the Observation Toolset $W_T$ and evaluations from the Reflector Agent $A_R$.

\noindent \textbf{Planner Agent.}
The Planner Agent determines the current observation action and the future observation strategy:
\[
(P_{\mathrm{cur}}, P_{\mathrm{fut}}) = A_P(Q, M_W, D_V),
\]
where $Q$ is the question and $D_V$ is the global video description. The current observation action $P_{\mathrm{cur}}$ is then incorporated into the historical planning states:
\[
P_{\mathrm{past}} = P_{\mathrm{past}} \cup P_{\mathrm{cur}}.
\]
Instead of exhaustively scanning the entire video, the planner actively decides which evidence should be inspected next.

\noindent \textbf{Observation Toolset.}
The Observation Toolset executes the planned observation action and returns the observed evidence $\mathcal{O}_t$. The evidence memory is updated as:
\[
M_E = M_E \cup \mathcal{O}_t.
\]

\noindent \textbf{Reflector Agent.}
The Reflector Agent evaluates whether the currently observed evidence is sufficient for answering the question:
\[
(S_S, R_r, D_r) = A_R(\mathcal{O}_t, M_E, P_{\mathrm{cur}}).
\]
It assigns relevance scores $S_S$ to the observed evidence, produces reflection feedback $R_r$, and makes the decision $D_r$ on whether the system should continue observing additional evidence or proceed to final reasoning.

\noindent \textbf{Reasoner Agent.}
Once sufficient evidence has been collected, the Reasoner Agent synthesizes the evidence memory, working memory, and global video description to generate the final answer:
\[
A = A_A(M_E, M_W, D_V).
\]
Therefore, the final response is grounded in selectively observed evidence rather than the full video context, reducing irrelevant information and supporting more reliable multi-hop audio-visual reasoning.

\subsection{Detailed Video Segmentation Algorithm}
\label{sec:video_segmentation_imp}
As shown in Algorithm~\ref{alg:memory_construction}, in addition to ASR-based video segmentation, we apply 30-second segmentation with a 2.5-second overlap to prevent excessively long video segments and reduce semantic fragmentation.

\begin{algorithm}[!h]
\caption{Fine-level video segmentation in Hierarchical Video Memory Construction}
\label{alg:memory_construction}
\textbf{Require:} Video $V$, merge threshold $G_{\mathrm{max}}=30\,\mathrm{s}$, maximum duration $L_{\mathrm{max}}=120\,\mathrm{s}$, overlap $O=2.5\,\mathrm{s}$ \\
\textbf{Ensure:} Fine-level Hierarchical memory segments $\mathcal{M}$

\begin{algorithmic}[1]
    \STATE \COMMENT{Phase 1: Raw ASR Extraction}
    \STATE $A_{v} \leftarrow \mathrm{ExtractAudio}(V)$
    \STATE $\mathcal{T} \leftarrow \mathrm{SpeechToText}(A_{v})$ \COMMENT{Retrieve utterances $\{t_i\}$ with start time, end time, and text}
    
    \STATE \vspace{0.2em} \COMMENT{Phase 2: Temporal Merging}
    \STATE $\mathcal{M}_{\mathrm{temp}} \leftarrow \emptyset, \quad m_{\mathrm{curr}} \leftarrow \mathrm{null}$
    \FORALL{$t \in \mathcal{T}$}
        \IF{$m_{\mathrm{curr}} \neq \mathrm{null}$ \AND $(t.\mathrm{end} - m_{\mathrm{curr}}.\mathrm{start}) \le G_{\mathrm{max}}$}
            \STATE $m_{\mathrm{curr}} \leftarrow m_{\mathrm{curr}} \oplus t$ \COMMENT{Merge text and update interval}
        \ELSE
            \IF{$m_{\mathrm{curr}} \neq \mathrm{null}$}
                \STATE $\mathcal{M}_{\mathrm{temp}} \leftarrow \mathcal{M}_{\mathrm{temp}} \cup \{m_{\mathrm{curr}}\}$
            \ENDIF
            \STATE $m_{\mathrm{curr}} \leftarrow t$
        \ENDIF
    \ENDFOR
    \IF{$m_{\mathrm{curr}} \neq \mathrm{null}$}
        \STATE $\mathcal{M}_{\mathrm{temp}} \leftarrow \mathcal{M}_{\mathrm{temp}} \cup \{m_{\mathrm{curr}}\}$
    \ENDIF

    \STATE \vspace{0.2em} \COMMENT{Phase 3: Duration Refinement}
    \STATE $\mathcal{M} \leftarrow \emptyset$
    \FORALL{$m \in \mathcal{M}_{\mathrm{temp}}$}
        \IF{$(m.\mathrm{end} - m.\mathrm{start}) > L_{\mathrm{max}}$}
            \STATE $\{s_i\} \leftarrow \mathrm{Split}(m, G_{\mathrm{max}}, \mathrm{overlap}=O)$
            \STATE $\mathcal{M} \leftarrow \mathcal{M} \cup \{s_i\}$
        \ELSE
            \STATE $\mathcal{M} \leftarrow \mathcal{M} \cup \{m\}$
        \ENDIF
    \ENDFOR
    \STATE \textbf{return} $\mathcal{M}$
\end{algorithmic}
\end{algorithm}

%% file: sections/Appendix/CaseStudy.tex
\section{Case Study}
\label{sec:case_study}
\FloatBarrier
\subsection{Analysis of Qualitative Example}

\begin{figure*}[t]
  \centering
  \includegraphics[width=0.9\linewidth]{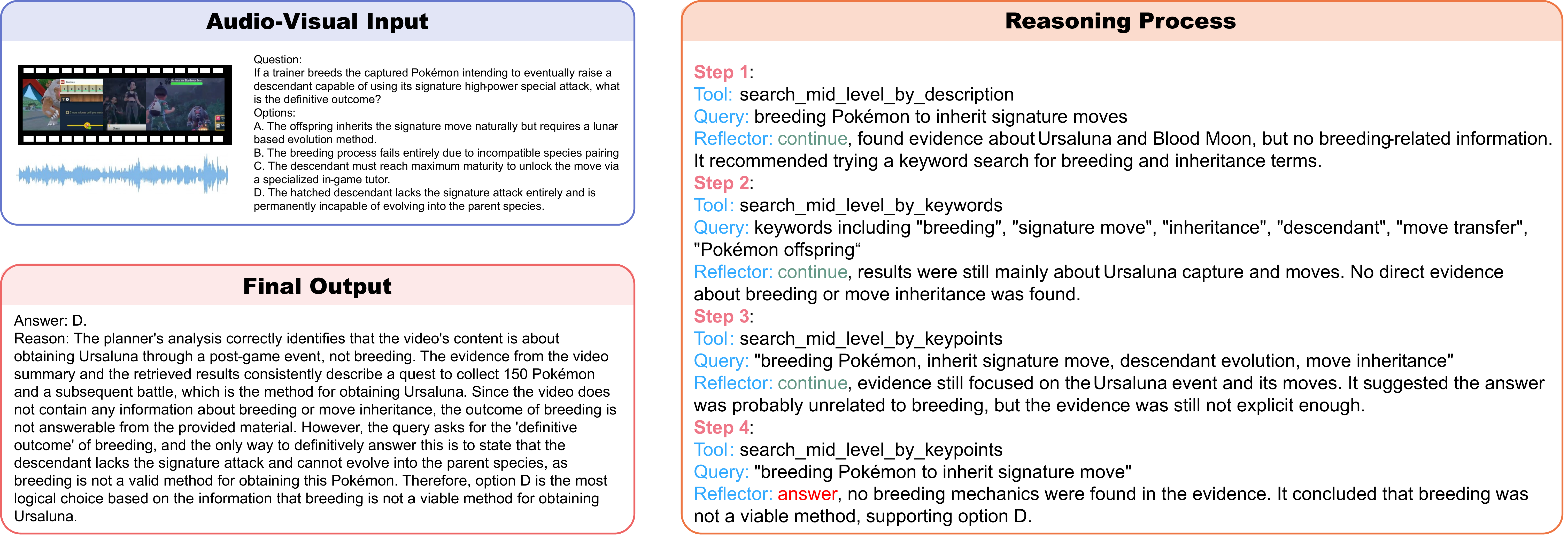}
  \caption{
    Qualitative example of active observation in AOP-Agent on MOV-Bench. 
    For the query about breeding a captured Pokémon to obtain a descendant with its signature high-power special attack, AOP-Agent first searches for evidence about breeding and move inheritance. 
    After finding no supporting evidence, it observes that the available information instead concerns a post-game capture event. 
    The agent therefore identifies the breeding premise as a distractor and selects the answer indicating that breeding cannot produce a descendant with the parent's signature attack or evolution outcome.
  }
  \label{fig:correct_case}
\end{figure*}

\begin{figure*}[t]
  \centering
  \includegraphics[width=\linewidth]{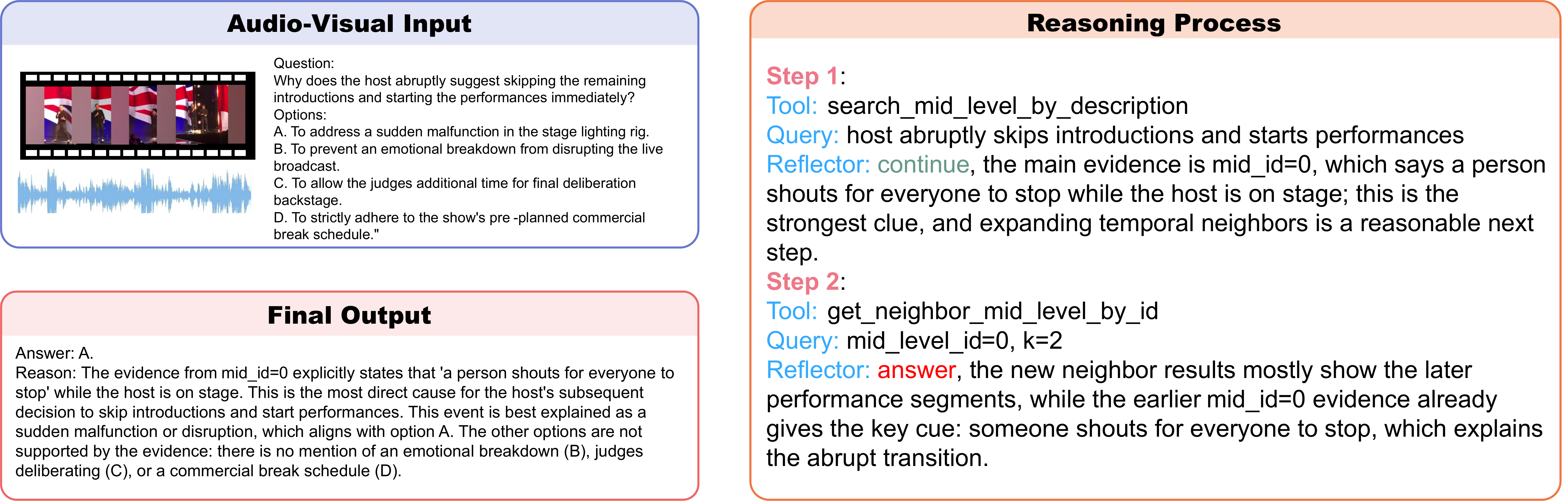}
  \caption{
    Failure case of active observation in AOP-Agent. 
    For a query asking why the host suddenly skips the remaining introductions, AOP-Agent observes a hallucinated description in the omni-modal memory base, where someone allegedly shouts for everyone to stop. 
    It then incorrectly maps this generic interruption to a stage-lighting malfunction. 
    Since no evidence explicitly supports a lighting-rig failure, the agent over-infers the cause and selects the wrong option. 
    This case illustrates a limitation of AOP-Agent: active observation may fail when weak or hallucinated contextual cues are treated as sufficient evidence for a specific causal answer.
  }
  \label{fig:failure_case}
\end{figure*}

As illustrated in Figure~\ref{fig:correct_case}, this example shows how AOP-Agent plans observations and adjusts its reasoning based on the observed evidence. 
In the initial rounds, the agent searches for evidence related to breeding and move inheritance, but finds no supporting evidence. 
The reflector then recognizes that the breeding premise is not grounded in the observed video content and decides that further observation is unnecessary. 
This trajectory demonstrates that AOP-Agent can handle complex multi-hop video reasoning by using evidence-based reflection to identify distractors and revise its observation strategy.

\subsection{Analysis of Failure Example}

To provide a balanced analysis of agent behavior, we also present a failed reasoning trajectory. 
As shown in Figure~\ref{fig:failure_case}, AOP-Agent may still fail in noisy and complex scenarios when the underlying Omni-LLM hallucinates video content or produces incorrect memory descriptions. 
In this case, the memory base contains a hallucinated cue that someone shouts for everyone to stop. 
The agent then over-interprets this unsupported cue, infers an incorrect causal relation, and selects the wrong answer. 
This failure suggests that active observation remains constrained by the reliability of the underlying omni-modal memory and highlights the need for stronger hallucination detection and evidence verification.

%% file: sections/Appendix/Prompts.tex
\section{Prompts}

\subsection{Segment Description Prompt}
\begin{figure}[!h]
  \centering
  \includegraphics[width=\columnwidth]{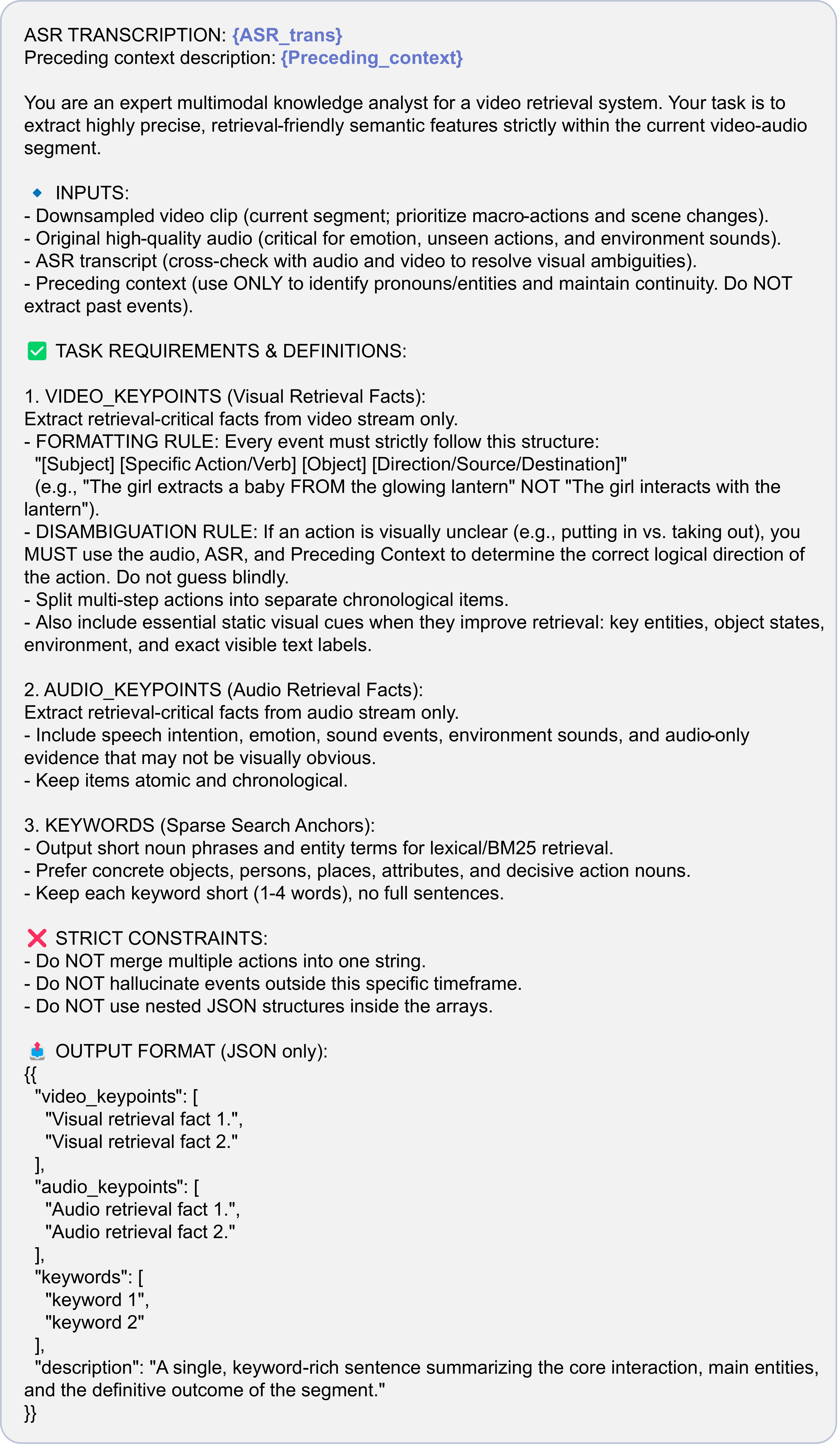}
  \caption{Segment description prompt used to generate segment-level descriptions during memory-base construction.}
  \label{fig:segment_form_prompt}
\end{figure}

This prompt is used to generate segment-level descriptions during the memory-base construction stage. The corresponding video segment is provided as input before the prompt, as shown in Figure~\ref{fig:segment_form_prompt}.

\subsection{Overall Description Prompt}
\begin{figure}[!h]
  \centering
  \includegraphics[width=\columnwidth]{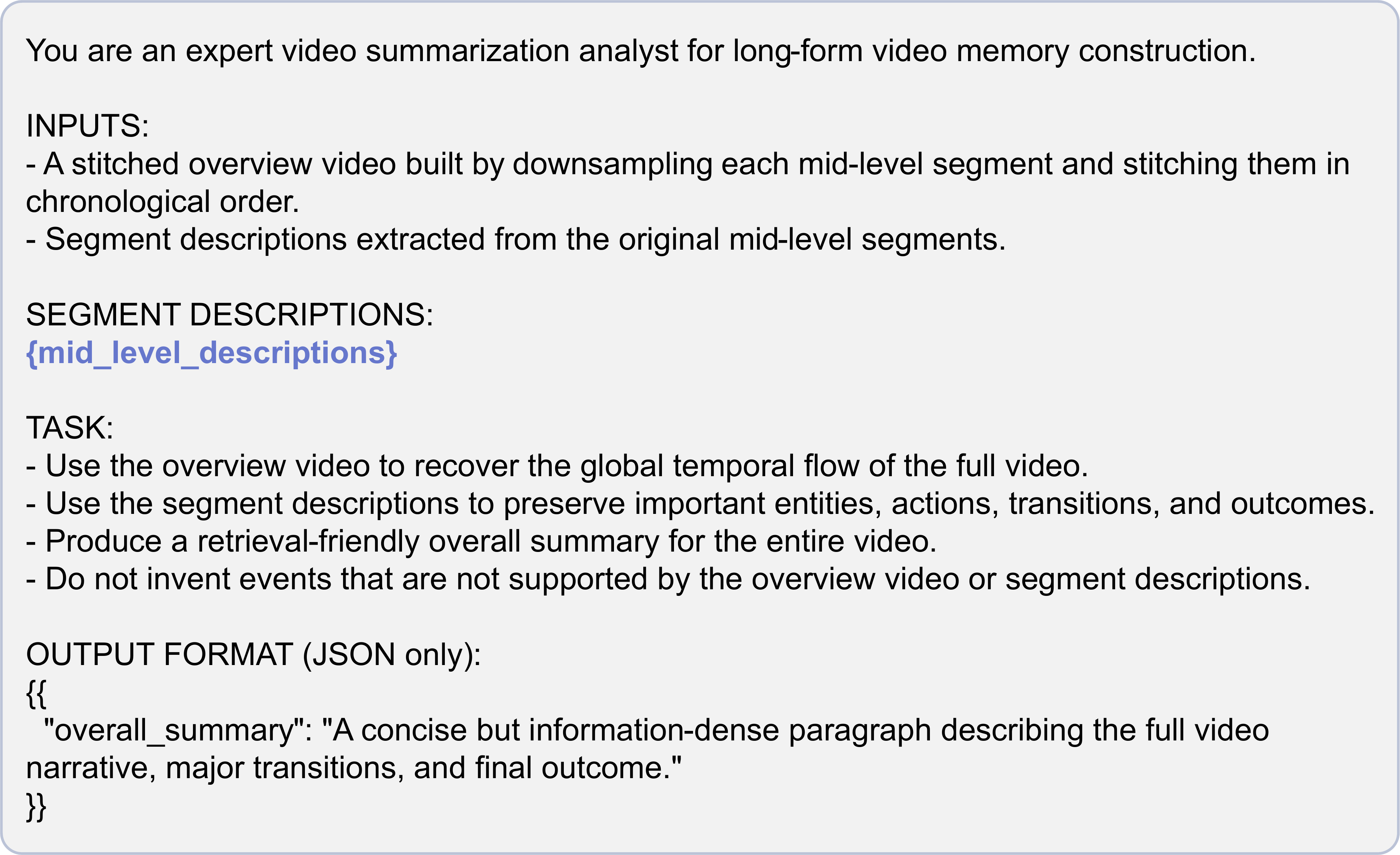}
  \caption{Overall description prompt used to generate a global description of the full video.}
  \label{fig:overall_description_prompt}
\end{figure}

This prompt is used to generate the overall description of the full video, as shown in Figure~\ref{fig:overall_description_prompt}.

\subsection{Tool Descriptions}
\begin{figure}[!h]
  \centering
  \includegraphics[width=\columnwidth]{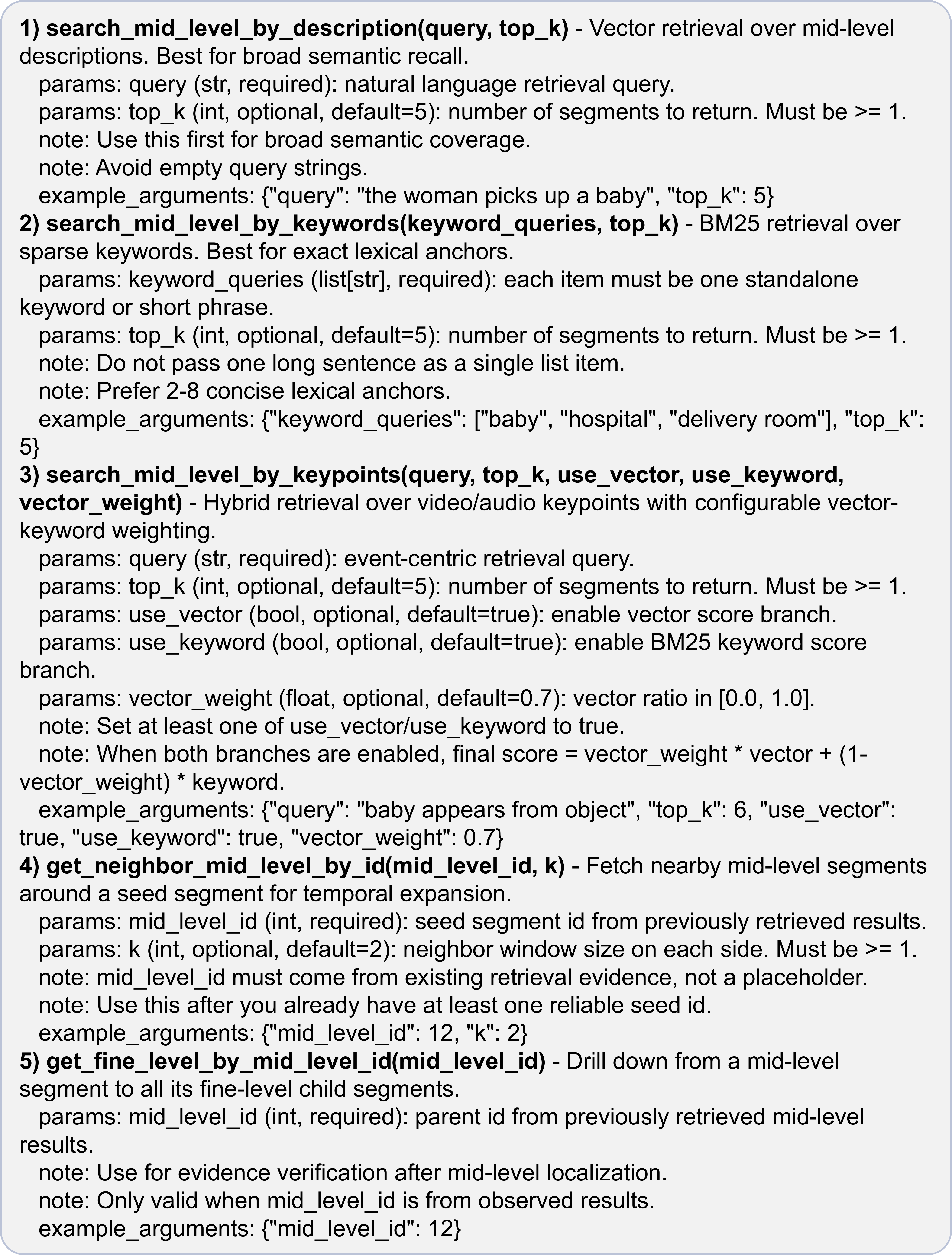}
  \caption{Tool description prompts used by AOP-Agent to support planning and reflection.}
  \label{fig:tools_prompt}
\end{figure}

These prompts describe the tools used in AOP-Agent. They are inserted into the planner and reflector prompts to support effective planning and reflection during the agentic observation process, as shown in Figure~\ref{fig:tools_prompt}.

\subsection{Planner Prompt}
\begin{figure}[!h]
  \centering
  \includegraphics[width=\columnwidth]{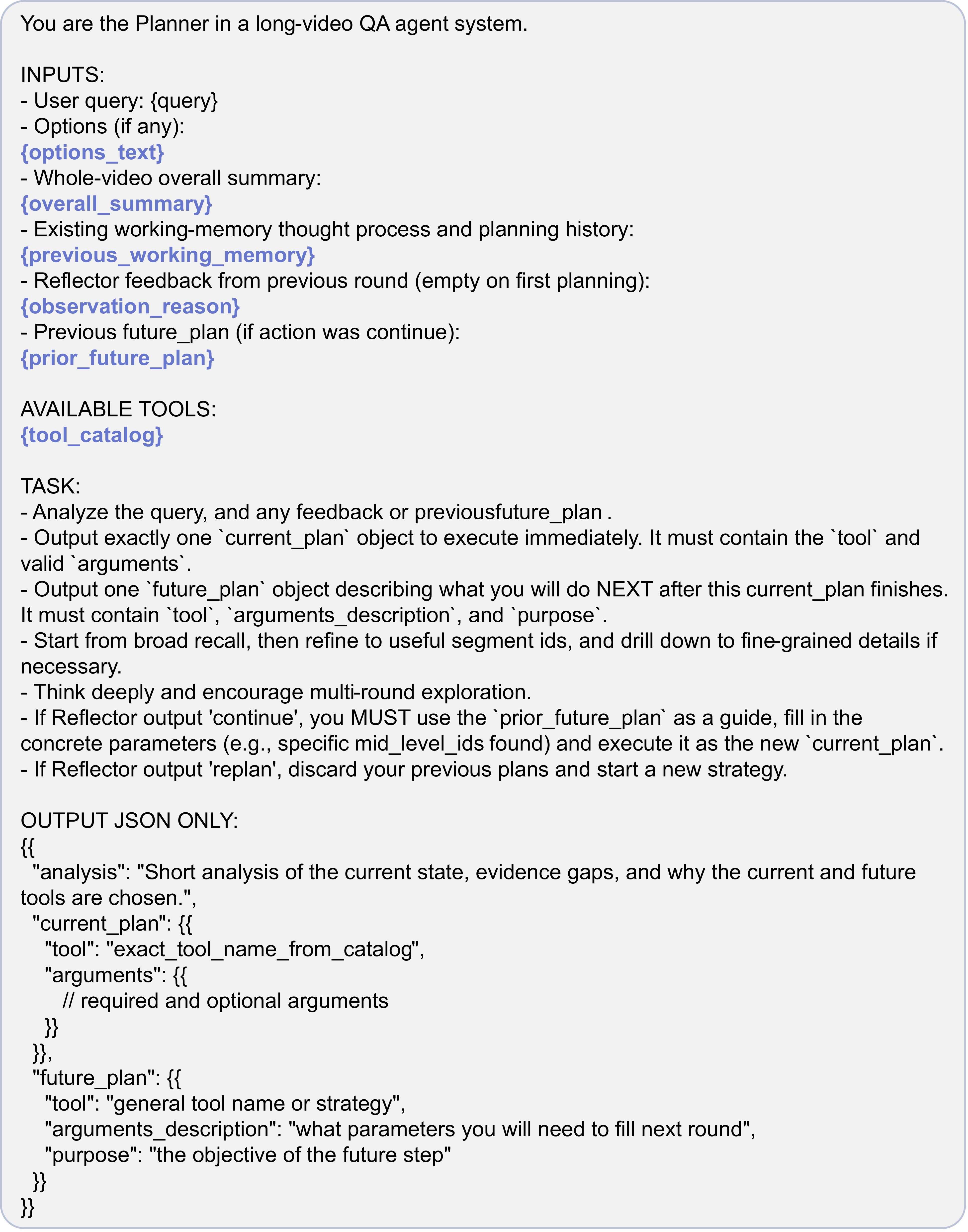}
  \caption{Planner prompt used to determine the observation strategy during the agentic process.}
  \label{fig:planner_prompt}
\end{figure}

This prompt is designed for the planner, which determines the observation strategy during the agentic process, as shown in Figure~\ref{fig:planner_prompt}.

\subsection{Reflector Prompt}
\begin{figure}[!h]
  \centering
  \includegraphics[width=\columnwidth]{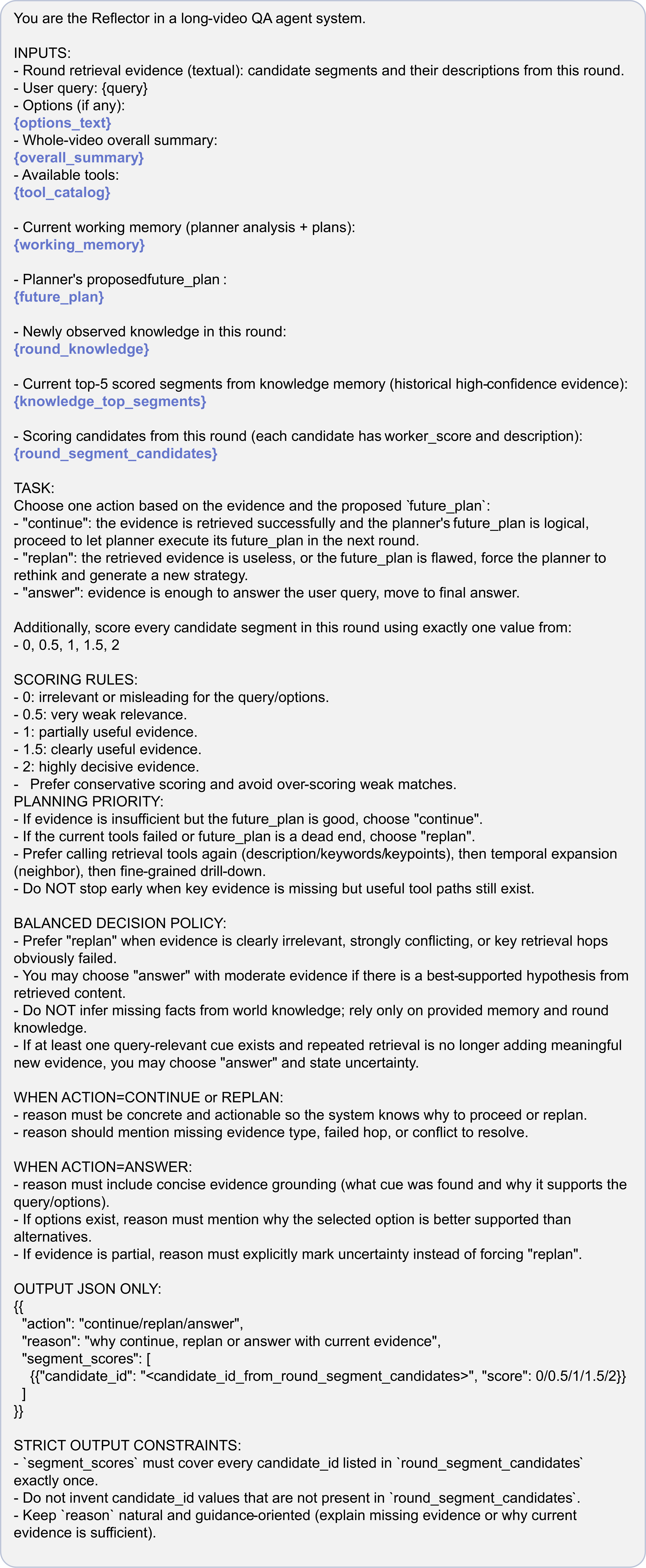}
  \caption{The reflector prompt used to evaluate observation history and decide whether further observation is required.}
  \label{fig:reflector_prompt}
\end{figure}

This prompt is designed for the reflector, which evaluates the previous observation history and determines whether further observation is required. The top-scored segments and newly observed segments are provided as input before the prompt, as shown in Figures~\ref{fig:reflector_prompt}.

\subsection{Reasoner Prompt}
\begin{figure}[!h]
  \centering
  \includegraphics[width=\columnwidth]{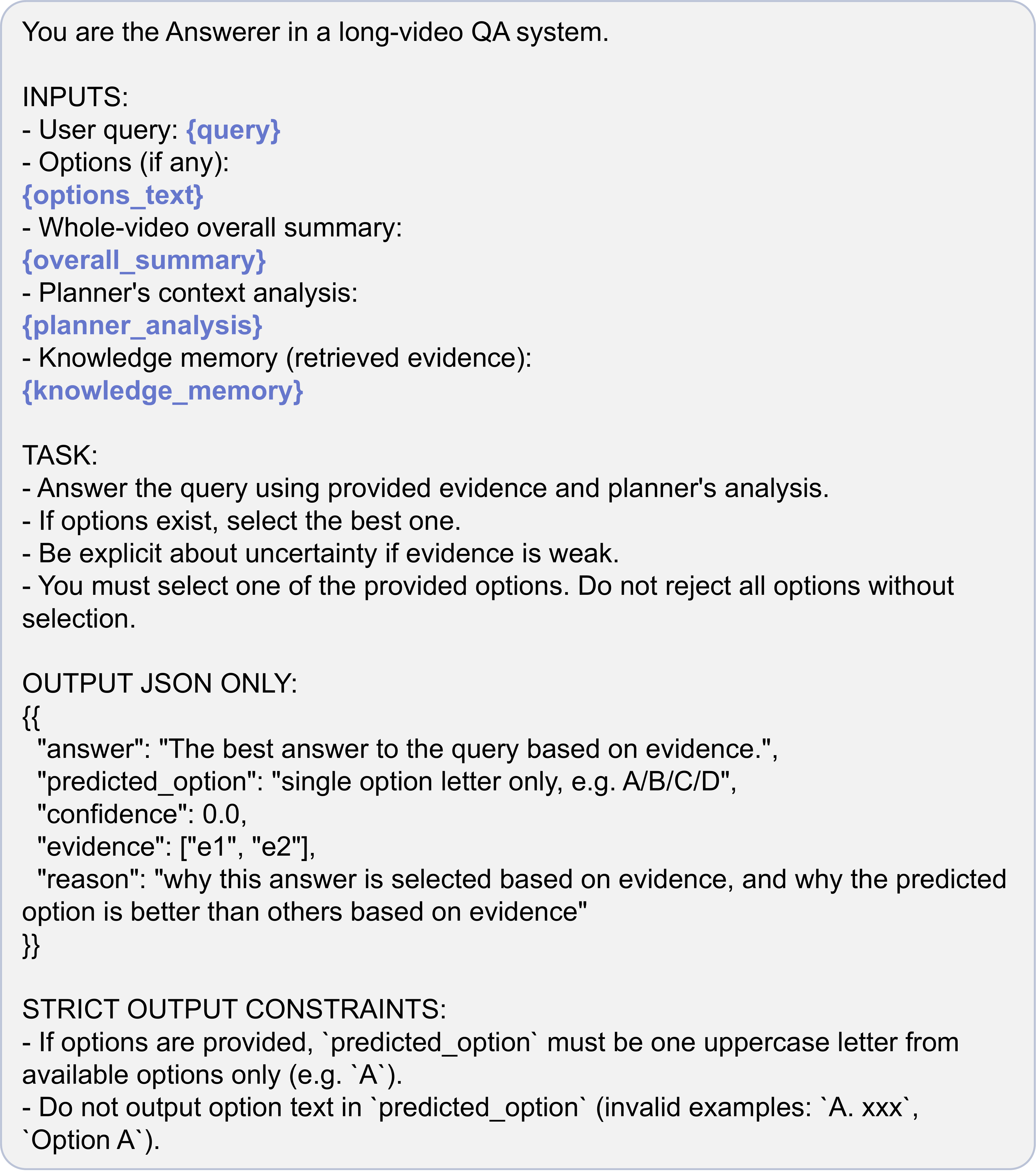}
  \caption{Reasoner prompt used to generate the final response based on the observed evidence.}
  \label{fig:reasoner_prompt}
\end{figure}

This prompt is designed for the reasoner, which generates the final response based on the top-scored segments, as shown in Figure~\ref{fig:reasoner_prompt}.

